%% file: main.tex
\documentclass[letterpaper]{article} % DO NOT CHANGE THIS
\usepackage[preprint]{aaai2027}  % Public preprint; author details will be filled later.
\usepackage[hyphens]{url}  % DO NOT CHANGE THIS
\usepackage{graphicx} % DO NOT CHANGE THIS
\usepackage{natbib}  % DO NOT CHANGE THIS AND DO NOT ADD ANY OPTIONS TO IT
\usepackage{caption} % DO NOT CHANGE THIS AND DO NOT ADD ANY OPTIONS TO IT
\usepackage{subcaption}
\usepackage{algorithm}
\usepackage{algorithmic}
\usepackage{booktabs}
\usepackage{amsmath}
\usepackage{amssymb}
\usepackage{multirow}
\usepackage{colortbl}   % rowcolor / cellcolor (xcolor already loaded by aaai2027.sty)
\usepackage{makecell}
\definecolor{cb}{RGB}{232,241,250}

\input{appendix/appendix_preamble}

\title{BiGrounder: Bidirectional Training Unlocks the Potential of GUI Grounding}
\author{
    \normalfont\normalsize
    Dengyao~Shi\textsuperscript{1,3,*},
    Tianyu~Fu\textsuperscript{2,*},
    Fei~Hu\textsuperscript{2,*},
    Anyang~Su\textsuperscript{2},
    Hanning~Wang\textsuperscript{2},\\
    Shuo~Wang\textsuperscript{2},
    Minghui~Wu\textsuperscript{2},
    Chenxu~Zhao\textsuperscript{2},
    and Zhen~Lei\textsuperscript{1}
}

\affiliations{
    \normalfont\normalsize
    \textsuperscript{1}MAIS, Institute of Automation, Chinese Academy of Sciences\\
    \textsuperscript{2}Mininglamp Technology\\
    \textsuperscript{3}School of Advanced Interdisciplinary Sciences,\\ University of Chinese Academy of Sciences\\
    \textsuperscript{1}\{shidengyao2026@ia.ac.cn, zhen.lei@ia.ac.cn\}
    \textsuperscript{*}Equal contribution.
}

\begin{document}

\maketitle
\pagestyle{plain}
\thispagestyle{plain}

\begin{abstract}
Graphical User Interface (GUI) grounding aims to map natural language instructions to precise operational coordinates, serving as a critical task for autonomous agent interaction. Current approaches formalize GUI grounding as predicting action-coordinate pairs from instructions, with end-to-end training supervised solely by final coordinates. Training datasets typically provide only a small number of descriptions per target coordinate, whereas real-world scenarios involve a much larger set of varied descriptions for the same target, which limits robustness to linguistic variation. While recent GUI grounding paradigms have begun employing reinforcement learning to enhance localization accuracy, models with limited exploratory capability before reinforcement yield constrained improvements even with identical training data during the reinforcement phase. To overcome these limitations, we propose a bidirectional reward paradigm that enables model self-reinforcement. During SFT, bidirectional training equips the model to generate action descriptions from target coordinates. During RL, the model samples semantically consistent instruction variations on-policy and grounds each variation back to its target. A cycle-consistency gate couples semantic quality to grounding success, turning the model's own instruction policy into a source of continually refreshed reinforcement trajectories. Our 4B-parameter model achieves competitive accuracy across five grounding benchmarks in our evaluation, indicating the effectiveness of the proposed training paradigm. https://github.com/Spell-shi/BiGrounder
\end{abstract}

% TODO: reinsert Figure 1 (overall teaser: figures/figure1.pdf) --- redraw for two-column layout.

\begin{figure*}[t]
    \centering
    \includegraphics[width=0.99\textwidth]{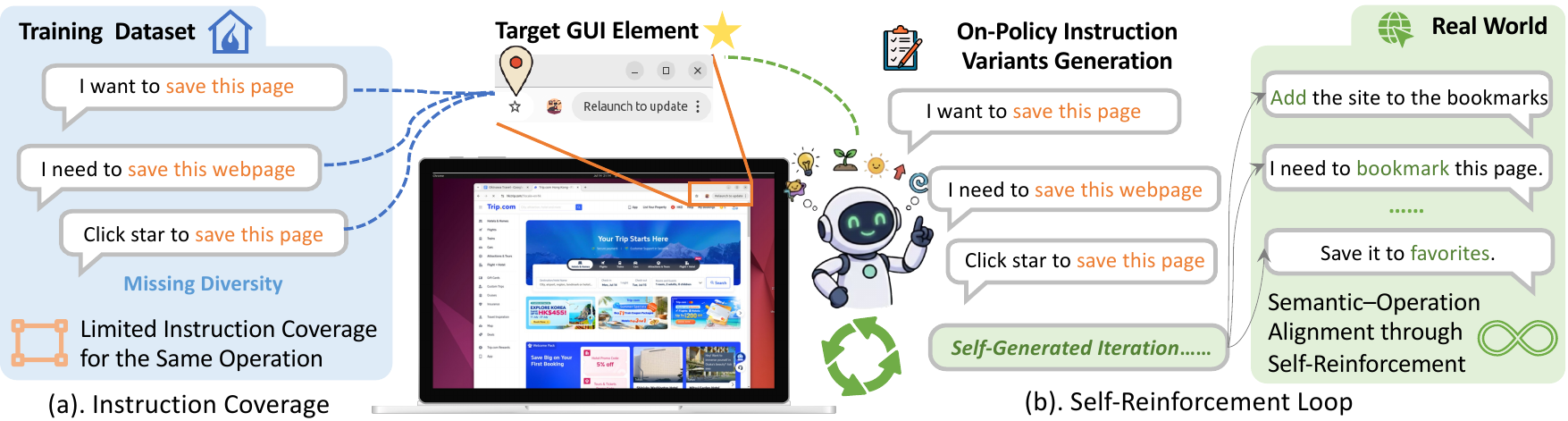}
    % \caption{Existing datasets \textbf{(a)} suffer from limited instruction coverage, whereas our bidirectional reward \textbf{(b)} enables semantic--operation alignment via self-reinforcement.}
    \caption{Existing datasets \textbf{(a)} suffer from limited instruction coverage, whereas our bidirectional reinforcement framework \textbf{(b)} samples on-policy instruction variations for the same UI element, strengthening semantic–element understanding.}
    \label{fig:overall_results}
\end{figure*}

\input{sections/introduction}
\input{sections/related_works}

\begin{figure*}[!t]
    \centering
    \includegraphics[width=0.99\textwidth]{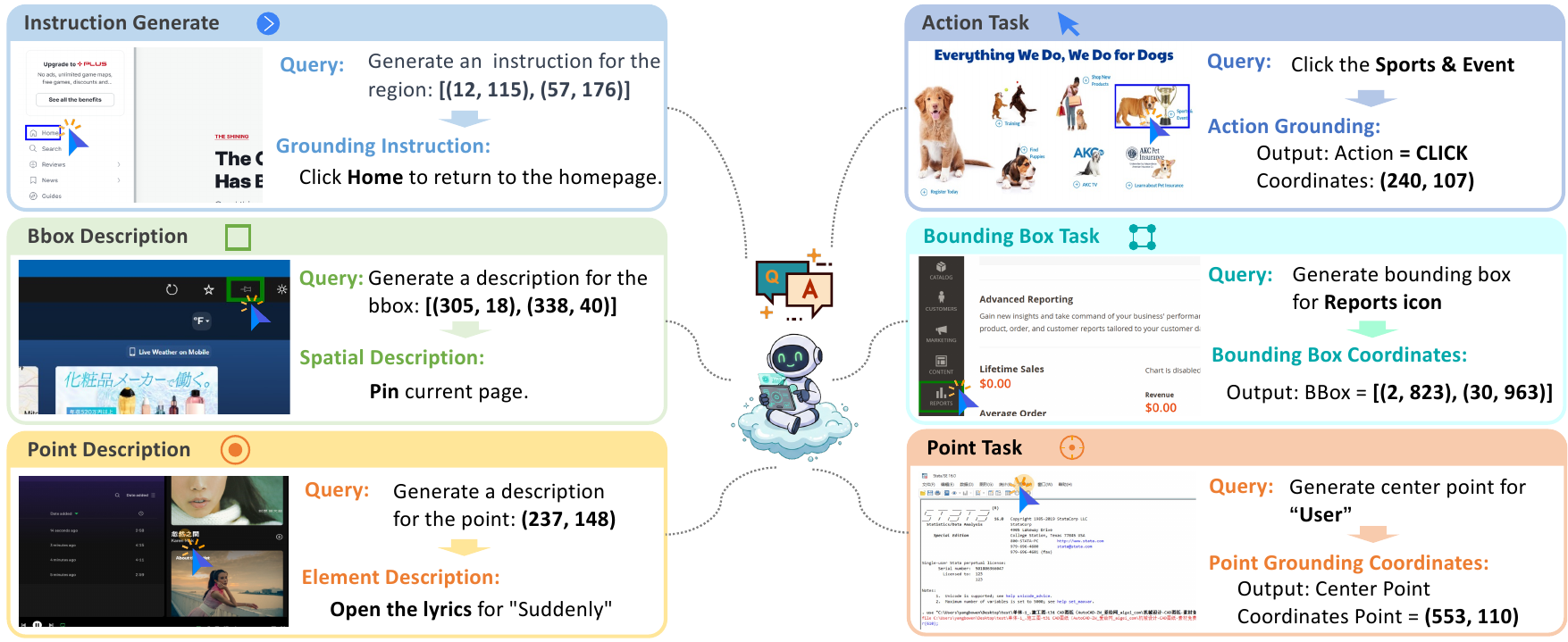}
    \caption{During the SFT stage, we organize the training data into six tasks: instruction generation, instruction grounding, bounding-box description generation, bounding-box localization, point-based description generation, and point localization.}
    \label{fig:data-0304}
\end{figure*}

\input{sections/method}

% \begin{figure*}[!t]
%     \centering
%     \includegraphics[width=0.99\textwidth]{images/rl-0305.pdf}
%     \caption{In the reinforcement learning phase, training is divided into two stages: a backward stage that obtains semantic rewards via instruction generation, and a forward stage that obtains grounding rewards.}
%     \label{fig:RL_loop}
% \end{figure*}

\input{sections/experiments}
\input{sections/ablation}
\input{sections/conclusion}

% ---- Bibliography ----
% aaai2027.sty automatically sets \bibliographystyle{aaai2027}; do NOT add one.
\bibliography{main}

% Flush the main text and its references before the appendix.
\clearpage
\input{appendix/appendix}

\end{document}

%% file: appendix/appendix_preamble.tex
\usepackage{graphicx}
\usepackage{natbib}
\usepackage{caption}
\usepackage{xcolor}

\usepackage{booktabs}
\usepackage{amsmath}    % equation*, cases, \wedge in the reward-hacking section
\usepackage{amssymb}    % \mathbb for the cycle-gate indicator
\newtheorem{samplingproposition}{Proposition}[section]
\usepackage{multirow}
\usepackage{makecell}
\usepackage{tabularx}  % fixed-width cells keep row shading continuous
\usepackage{colortbl}   % rowcolor / cellcolor (xcolor is already loaded by aaai2027.sty)
\usepackage{subcaption} % side-by-side subfigures
\usepackage{tcolorbox}  % prompt boxes in the dataset-construction section
\tcbuselibrary{breakable}
\usepackage{needspace} % keep prompt headings with their opening content
\usepackage{seqsplit}   % allow long no-space code tokens to wrap inside prompt boxes
\usepackage{cuted}      % strip env: full-width, page-breakable single-column-spanning block

\providecommand{\h}[1]{\textbf{\footnotesize \textsf{#1}}}
\definecolor{cb}{RGB}{232,241,250}

\definecolor{ratioTwenty}{HTML}{F7FBFE}
\definecolor{ratioThirty}{HTML}{F2F7FB}
\definecolor{ratioForty}{HTML}{EDF4FA}
\definecolor{ratioFifty}{HTML}{E7F0F8}
\usepackage{pifont}
\providecommand{\xmark}{\ding{55}}
\providecommand{\checkmark}{\ding{51}}

%% file: sections/introduction.tex
\section{Introduction}
\label{sec:intro}
Current approaches in GUI grounding predominantly focus on learning direct mappings from singular operational descriptions to screen coordinates. However, this paradigm overlooks a critical characteristic of real-world applications: the complex many-to-one mapping relationship between user instructions and target coordinates. Specifically, users may employ diverse expressions, varying levels of detail, and different linguistic styles to refer to the same interface element. Such variability in natural language can significantly impair coordinate prediction, especially when instructions differ in length or style.

To mitigate this challenge, existing research primarily adopts offline data augmentation strategies~\cite{UI-E2I-Synth,uground,AutoGUI,liu2025scalecua}, using instruction paraphrasing and expansion to improve model generalization. However, because the augmented instructions are generated in advance and remain fixed throughout training, the model can explore only a predefined sampling space that may not capture the breadth and evolving nature of instructional variations encountered in real-world scenarios.

From a training paradigm perspective, recent work has increasingly explored reinforcement learning~\cite{ui_r1,GUIG1} for improving GUI grounding performance, typically utilizing coordinate hits within target bounding boxes as the primary reward signal. Nevertheless, current reinforcement learning methods~\cite{SE_GUI,liu2025infiguig1} predominantly operate on fixed offline datasets, resulting in limited exploration capabilities.

To address the aforementioned limitations, we aim for agents to not only fit the "instruction$\rightarrow$coordinate" outcome mapping but also learn robust semantic-spatial alignment representations based on local visual evidence from coordinate-corresponding regions, thereby better adapting to diverse interfaces and instruction expressions in real-world scenarios. To this end, we propose BiGrounder: a reinforcement learning framework based on bidirectional semantic verification. Beyond traditional forward localization, the model generates instructions from annotated target coordinates and then grounds its own on-policy generations back to the same elements. This turns reverse generation into an endogenous exploration mechanism: as the policy evolves, it continually supplies instruction variations that are semantically anchored to the target yet reflect the model's current behavior. The resulting reverse and forward trajectories reinforce semantic--spatial alignment through cycle-consistent rewards. Experimental results demonstrate competitive results across multiple industry-standard benchmarks.

Our main contributions are summarized as follows:
\begin{itemize}
    \item \textbf{Constructing GUI understanding data from grounding annotations.} We reverse existing instruction-to-coordinate grounding samples into coordinate-conditioned instruction-generation data and incorporate them into bidirectional SFT, providing explicit supervision for learning the semantic and spatial properties of GUI elements.
    \item \textbf{On-policy bidirectional self-reinforcement. } We propose a cycle-consistent RL paradigm that uses the model's reverse policy to generate instruction variations on-policy and grounds them through the forward policy. This self-reinforcement mechanism can be directly integrated into standard SFT+RL pipelines.
    \item \textbf{Competitive performance on cross-platform benchmarks.} Experiments demonstrate that our model achieves competitive performance across five grounding benchmarks, including ScreenSpot-Pro~\cite{li2025screenspotpro}, ScreenSpot-V2~\cite{osaltas_and_screenspot_v2}, UI-Vision~\cite{ui_vision}, MMBench-GUI L2~\cite{mmbenchgui}, and OSWorld-G-R~\cite{jedi}.
\end{itemize}

%% file: sections/related_works.tex
\section{Related Works}

\subsection{High-Quality Dataset Construction and Instruction Augmentation}
Several works improve data quality and instruction coverage. UI-E2I-Synth~\cite{UI-E2I-Synth}, AutoGUI~\cite{AutoGUI}, OS-ATLAS~\cite{osaltas_and_screenspot_v2}, UGround~\cite{uground}, and JEDI~\cite{jedi} offline expand grounding datasets by generating diverse instructions and functionality descriptions with large language models, while GroundCUA~\cite{feizi2025groundingcomputeruseagents} and ScaleCUA~\cite{liu2025scalecua} highlight the value of high-quality human-annotated and GUI-understanding data. However, such offline augmentation confines instructions to predefined sample sets and cannot progressively evolve with the model during training, making it hard to cover the wide range of real-world instruction expressions.

\subsection{Reinforcement Learning in GUI Agents}
Recent research explores various strategies to enhance GUI grounding with RL. One line refines the reward signal: UI-R1~\cite{ui_r1} uses a composite accuracy/action/format reward and shows explicit \texttt{<think>} tokens are redundant, GUI-G1~\cite{GUIG1} curbs reward hacking by shaping reward granularity with hits, IoU, and element size, and GUI-G$^2$~\cite{GUI_G2} models elements as continuous Gaussian distributions to turn sparse binary hits into dense optimization. A second line targets the training framework and exploration: GUI-R1~\cite{luo2025guir1generalistr1style} adopts an RFT framework with a unified action space, SE-GUI~\cite{SE_GUI} filters overly difficult samples via attention foci from the previous round, infiGUI-G1~\cite{liu2025infiguig1} performs adaptive exploration policy optimization over multiple sampled answers, UI-Ins~\cite{chen2025uiinsenhancingguigrounding} alleviates policy collapse through instruction refinement and multi-view SFT, and GUI-RCPO~\cite{du2025testtimereinforcementlearninggui} self-evolves on unlabeled data via test-time sampling. Overall, RL is an effective and efficient approach for advancing GUI grounding. However, existing methods reward the model mainly by grounding accuracy and lack semantic rewards for generating correct instructions or descriptions of UI elements.

%% file: sections/method.tex
\section{Method}

\paragraph{\textbf{Overview.}}
We propose BiGrounder, a two-stage framework based on bidirectional semantic verification for enhancing visual-language alignment in GUI agents. Unlike traditional paradigms that only learn unidirectional "instruction$\rightarrow$coordinate" mappings, BiGrounder additionally learns a reverse generation policy conditioned on ground-truth target coordinates. During SFT, the model jointly learns forward localization and reverse description, acquiring both the ability to locate elements from instructions and to generate semantic descriptions from element regions. During RL, this reverse capability becomes an on-policy source of instruction variation. The cycle comprises two steps: (1) \emph{instruction generation}: conditioned on the ground-truth target coordinate $o^*$, the policy generates descriptions or interaction instructions; and (2) \emph{instruction grounding}: the same policy grounds each generated instruction, and the resulting coordinate is evaluated against $o^*$.

\subsection{Bidirectional Task Construction Method}
To enable the model to simultaneously possess GUI Grounding and self-generated user instruction capabilities, we design bidirectional training tasks: predicting coordinates from user instructions and generating user instructions from coordinates. As shown in Fig.~\ref{fig:data-0304}, the GUI Grounding task aims to map natural language instructions to specific interaction operations on graphical user interfaces. The model generates structured sequences containing action types and screen coordinates to explicitly indicate the location of target elements, thereby achieving alignment between visual features and precise spatial coordinates. The GUI Description task generates unstructured natural language sequences based on prompts and interface inputs to describe the elements corresponding to specific coordinates or the operations on these elements. Since each element is paired with multiple descriptions during bidirectional training, the model can sample varied, semantically consistent instructions in the subsequent RL stage.

\subsection{Model Training - Supervised Fine-tuning Stage}
By unifying strictly structured outputs (actions/coordinates) and unstructured semantic outputs (descriptions) into a single generation framework, the model learns to align visual features simultaneously with "functional semantics" and "precise spatial coordinates." The optimization objective is defined as:
\begin{equation}
    \mathcal{L} = - \mathbb{E}_{(I, x, Y) \sim \mathcal{D}} \left[ \sum_{t=1}^{T} \log P(y_t \mid y_{<t}, x, I) \right]
\end{equation}
where $\mathcal{D}$ denotes the multi-task training dataset. For a target sequence $Y$ of length $T$, $y_t$ represents the token at step $t$, conditioned on the input image $I$, the text instruction $x$, and the preceding tokens $y_{<t}$ under teacher forcing.
The supervised fine-tuning stage utilizes bidirectional data simultaneously, enabling the model to learn UI operations while comprehending the meaning of UI elements, thereby laying the foundation for subsequent reinforcement learning.

\begin{figure*}[!t]
    \centering
    \includegraphics[width=0.99\textwidth]{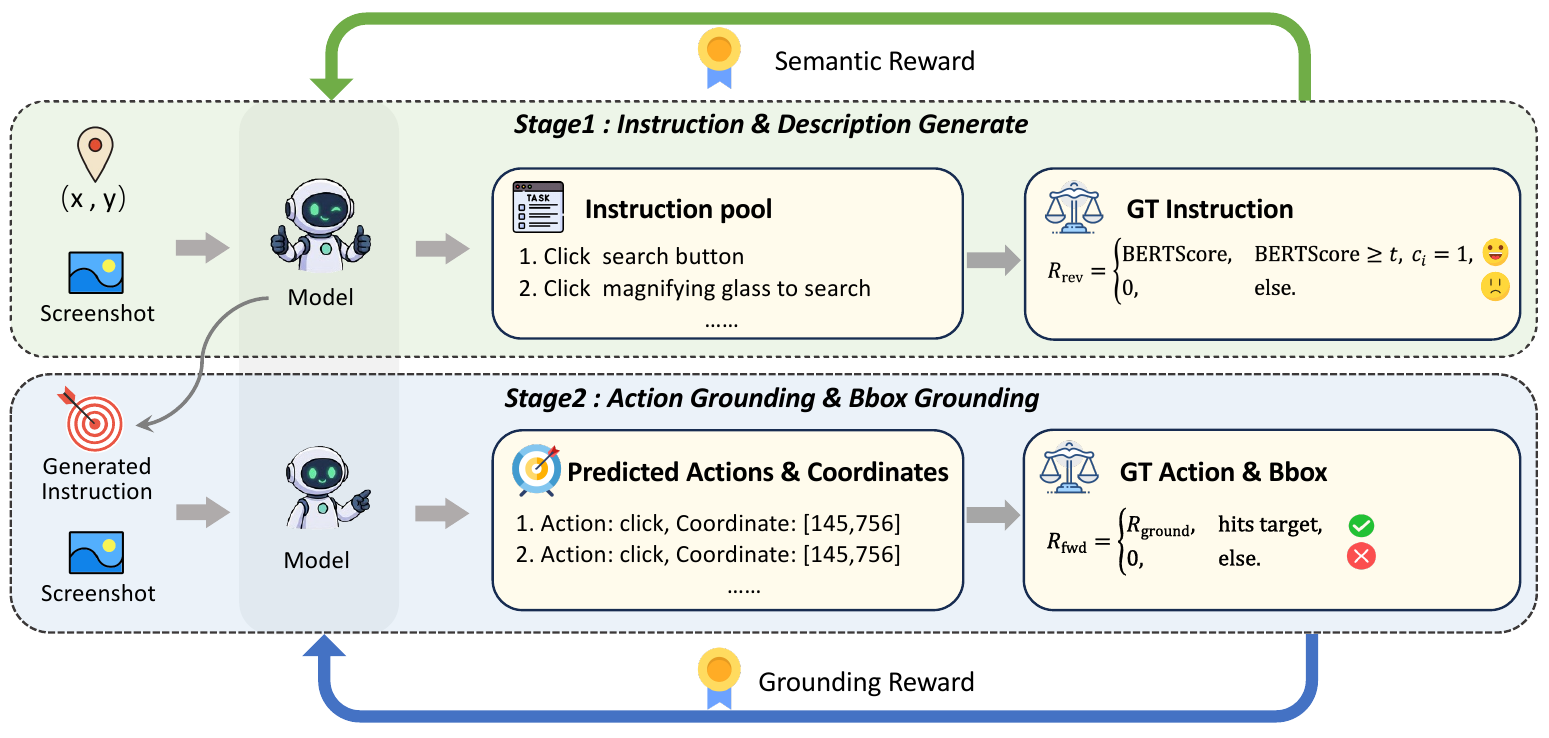}
    \caption{The reinforcement stage couples on-policy instruction-generation trajectories with their corresponding instruction-grounding trajectories.}
    \label{fig:RL_loop}
\end{figure*}

\subsection{Model Training - Reinforcement Stage}
\paragraph{\textbf{Reinforcement Sample Selection}}
We utilize the SFT model to construct the RL dataset via rejection sampling with an inference temperature of 0.9. A sample is retained only if it satisfies two criteria:
(1) Semantic Consistency. Among the four backward-generated instructions, at least three achieve a BERTScore greater than 0.85 with the original instruction.
(2) Difficulty Constraint. When performing forward inference with these four instructions, the model must succeed at least once but no more than three times (i.e., the number of correct predictions $s \in [1,3]$), ensuring the selection of non-trivial yet solvable samples

\begin{algorithm}[t]
\caption{Bidirectional Cycle-Consistent RL with GRPO}
\label{alg:cycle_rl}
\providecolor{stageBlue}{RGB}{31,78,121}
\providecolor{commentGray}{RGB}{96,96,96}
\begin{algorithmic}[1]
\renewcommand{\algorithmiccomment}[1]{\hfill\textcolor{commentGray}{\textit{// #1}}}
\REQUIRE Dataset $\mathcal{D}$, policy $\pi_\theta$, reference $\pi_{ref}$
\REQUIRE $G=4$, $K=4$, semantic threshold $t=0.85$
\WHILE{not converged}
    \STATE Sample batch $\mathcal{B} \sim \mathcal{D}$
    \FOR{each $(s, o^*, x) \in \mathcal{B}$}
        \STATE \textcolor{stageBlue}{\textbf{Stage 1: Instruction Generation}}
        \STATE Sample $G$ trajectories $\{\hat{x}_i\}_{i=1}^{G} \sim \pi_\theta(\hat{x}\mid s,o^*)$
        \FOR{$i = 1$ to $G$}
            \STATE $b_i \leftarrow \operatorname{BERTScore}(\hat{x}_i,x)$
            \STATE $q_i \leftarrow \hat{x}_i$ if $b_i \ge t$, else $x$ \COMMENT{GT fallback}
            \STATE \textcolor{stageBlue}{\textbf{Stage 2: Instruction Grounding}}
            \STATE Sample $K$ trajectories $\{\tau_{i,k}\}_{k=1}^{K} \sim \pi_\theta(\tau\mid s,q_i)$
            \STATE $r_{i,k}^{fwd} \leftarrow R_{\mathrm{ground}}(\tau_{i,k},o^*)$ \COMMENT{Grounding reward}
            \STATE $c_i \leftarrow \mathbb{I}[\exists k:r_{i,k}^{fwd} \ge 0.9]$ \COMMENT{Cycle gate}
            \STATE $r_i^{rev} \leftarrow R_{\mathrm{rev}}(b_i,c_i)$ \COMMENT{Semantic reward}
        \ENDFOR
        \STATE $A^{fwd}_{i,:} \leftarrow \operatorname{GRPOAdv}(r^{fwd}_{i,:})$, $i=1,\ldots,G$
        \STATE $A^{rev} \leftarrow \operatorname{GRPOAdv}(r^{rev})$
        \STATE Compute per-trajectory losses $\ell^{rev}$ and $\ell^{fwd}$
        \STATE $\mathcal{L}_{Bi} \leftarrow \operatorname{SeqMean}(\ell^{rev} \cup \ell^{fwd})$ \COMMENT{seq-mean-token-mean}
        \STATE Update $\pi_\theta$ using $\mathcal{L}_{Bi}$
    \ENDFOR
\ENDWHILE
\end{algorithmic}
\end{algorithm}

\paragraph{\textbf{Reverse Reward}}
In the reverse phase, our goal is to generate instructions that are semantically valid and operationally groundable. We employ Group Relative Policy Optimization to sample a set of candidate instructions $\{\hat{x}_i\}_{i=1}^{G}$ given the target coordinate. We use BERTScore~\cite{zhang2020bertscoreevaluatingtextgeneration} to measure candidate-to-reference consistency. Its precision, recall, and F-score are computed from token-embedding similarities as follows:

\begin{align}
    R_{\text{BERT}} &= \frac{1}{|x|} \sum_{x_i \in x} \max_{\hat{x}_j \in \hat{x}} \mathbf{x}_i^\top \hat{\mathbf{x}}_j \label{eq:bert_r} \\
    P_{\text{BERT}} &= \frac{1}{|\hat{x}|} \sum_{\hat{x}_j \in \hat{x}} \max_{x_i \in x} \mathbf{x}_i^\top \hat{\mathbf{x}}_j \label{eq:bert_p} \\
    F_{\text{BERT}} &= 2 \cdot \frac{P_{\text{BERT}} \cdot R_{\text{BERT}}}{P_{\text{BERT}} + R_{\text{BERT}}} \label{eq:bert_f}
\end{align}

Let $b_i=F_{\mathrm{BERT}}(\hat{x}_i,x)$ and $c_i=\mathbb{I}[\exists k:\operatorname{success}(\tau_{i,k},o^*)]$. A forward grounding trajectory is considered successful when its reward is at least $0.9$. We retain the semantic score only when the candidate passes the semantic threshold and completes cycle verification:
\begin{equation}
R_{\mathrm{rev}}=
\begin{cases}
\operatorname{BERTScore}(\hat{x}_i,x), & b_i\ge t,\ c_i=1,\\
0, & \text{otherwise}.
\end{cases}
\label{eq:reverse_reward}
\end{equation}
Here, $R_{\mathrm{rev}}$ denotes the reverse reward function, while $r_i^{\mathrm{rev}}$ is its value for the $i$-th instruction trajectory; $t=0.85$ is the semantic threshold. During training, we generate $G=4$ instructions and perform $K=4$ grounding rollouts for each instruction. If $b_i\ge t$, the generated instruction $\hat{x}_i$ enters its corresponding forward loop. Otherwise, we set $r_i^{\mathrm{rev}}=0$ and use the ground-truth instruction $x$ for that forward loop instead. For a qualified generation, its BERTScore reward is retained only if at least one of its $K$ grounding rollouts returns to the target. Thus, every instruction branch contributes forward grounding trajectories, while only semantically qualified and cycle-consistent generations update the reverse policy.

\paragraph{\textbf{Forward Reward}}
To cater to different grounding granularities, we use task-specific grounding rewards for action/point and region outputs:
\begin{equation}\label{eq:total_reward}
R_{\mathrm{ground}} =
\begin{cases}
R_{\mathcal{A}} + R_{\mathcal{C}} + R_{\mathcal{F}}, & \text{action grounding},\\
R_{\mathrm{point}}, & \text{point grounding},\\
R_{\mathrm{bbox}}, & \text{bounding box}.
\end{cases}
\end{equation}

\paragraph{\textbf{Action-Coordinate Alignment Reward:}}
For element interactions, $R_{\mathcal A}=0.1$ is awarded for the correct action type, and $R_{\mathcal C}=0.9$ is awarded when its predicted point falls inside the target bounding box; the coordinate reward is granted only if the action type is correct. Invalid output formats receive $R_{\mathcal F}=-0.2$, while valid formats receive no format penalty. For point grounding, no action type is involved, so we use a binary hit reward $R_{\mathrm{point}}=1$ when the predicted point falls inside the target bounding box and $0$ otherwise.

\paragraph{\textbf{BBox-based Reward:}}
For the bounding box prediction reward, we adopt the reward design of StepGUI~\cite{step-gui}. To ensure pixel-level precision, we implement a dense reward mechanism based on tolerance-normalized high-order decay. We define the normalized deviation as $\hat{\delta}_k = |k_{pred} - k_{gt}| / \tau_k$. For Region Grounding, we decouple the bounding box into center coordinates $(c_x, c_y)$ and dimensions $(w, h)$, using the same pixel tolerance for all four terms, $\tau_{c_x}=\tau_{c_y}=\tau_w=\tau_h=6$. We construct a composite geometric energy term $E_{geom}$, weighting down dimensional errors via the coefficient $\gamma$:
\begin{equation}
E_{g} = \hat{\delta}_{cx}^4 + \hat{\delta}_{cy}^4 + \gamma \hat{\delta}_w^4 + \gamma \hat{\delta}_h^4
\end{equation}
Here, we set $\gamma = 0.5$. To simultaneously optimize alignment precision and area overlap, the final bounding box reward $R_{bbox}$ is a weighted fusion of the geometric score and the Intersection over Union (IoU):
\begin{equation}
R_{bbox} = \alpha \cdot \exp(-E_{g}) + (1 - \alpha) \cdot \text{IoU}(b_{pred}, b_{gt})
\end{equation}
We set $\alpha = 0.8$, aiming to prioritize geometric centrality while retaining IoU as a regularization constraint for shape consistency.

Finally, we employ GRPO~\cite{grpo} to optimize the reverse-to-forward cyclic process. Within each trajectory group, the advantage is obtained by normalizing the rewards, $\hat{A}_{i} = (r_i - \mathrm{mean}(\mathbf{r})) / (\mathrm{std}(\mathbf{r})+\epsilon_{\mathrm{num}})$, where $\epsilon_{\mathrm{num}}=10^{-6}$. The per-trajectory clipped loss is
\begin{equation}
\ell_i(\theta)=-\min\!\left(\rho_i\hat{A}_i,\bar{\rho}_i\hat{A}_i\right)+\beta D_{\mathrm{KL}}\!\left(\pi_\theta\|\pi_{\mathrm{ref}}\right),
\end{equation}
where $\rho_i = \pi_{\theta}(o_i \mid I,S) / \pi_{\mathrm{old}}(o_i \mid I,S)$ is the importance ratio, $\bar{\rho}_i=\operatorname{clip}(\rho_i,1-\epsilon_{\mathrm{clip}},1+\epsilon_{\mathrm{clip}})$, and $\beta$ weights the KL penalty against the reference SFT model $\pi_{\mathrm{ref}}$. The $G$ reverse trajectories form one GRPO group, while the $K$ grounding rollouts associated with each generated instruction form a forward group; their advantages are normalized within the corresponding groups.

After group-wise advantage estimation, we jointly aggregate all trajectories using \textbf{seq-mean-token-mean} (each trajectory is averaged over its own tokens, then averaged with equal weight), so every trajectory contributes equally regardless of length:
\begin{equation}
\mathcal{L}_{\mathrm{Bi}}
= \frac{1}{G+GK}\left(
\sum_{i=1}^{G}\ell_i^{\mathrm{rev}}
+ \sum_{i=1}^{G}\sum_{k=1}^{K}\ell_{i,k}^{\mathrm{fwd}}
\right).
\label{eq:bi_objective}
\end{equation}
where $\ell_i^{\mathrm{rev}}$ and $\ell_{i,k}^{\mathrm{fwd}}$ are the reverse and forward instances of the loss in Eq.~(9). Under seq-mean-token-mean with $G=K=4$, the reverse and forward trajectories contribute $0.2$ and $0.8$ by construction.

%% file: sections/experiments.tex
\section{Experiments}
\subsection{Experimental Setup}
\paragraph{\textbf{Implementation details. }}
\input{tables/grounding_summary_backbone}
Model Architectures. Our primary model, BiGrounder-4B, is built on the Qwen3-VL-4B-Instruct backbone. To evaluate generalization across model scales and backbone families, we additionally validate BiGrounder on Qwen2.5-VL-3B-Instruct and Qwen2.5-VL-7B-Instruct. Following our ablation studies, we set the BERTScore threshold $t$ to 0.85 and all center and size tolerance thresholds to 6 pixels.

\begin{itemize}
    \item{SFT Stage. }
    We adopt the open-source grounding data from ScaleCUA~\cite{liu2025scalecua} and GroundCUA~\cite{feizi2025groundingcomputeruseagents}, and construct the corresponding reverse coordinate$\rightarrow$instruction data to add into training. For samples with only element descriptions, we employ Gemini 2.5-Flash to rewrite the descriptions into user instructions. We will release the rewritten instructions together with the exact prompts used. With a total of 4,482k samples, the training is conducted for 2 epochs with a global batch size of 640 and a learning rate of $1\times 10^{-5}$, aiming to endow the model with fundamental capabilities for both description generation and element grounding.

    \item{RL stage. }
    We use approximately 63k filtered samples for the RL stage. Unlike offline augmentation with its fixed sample set, our framework samples instruction variations on-policy from the current model at a temperature of 0.9. We employ GRPO for 1 epoch. For each sample, we generate 4 instructions in the backward phase, each followed by 4 grounding rollouts. We set the training batch size to 64 and the learning rate to $5\times 10^{-7}$.
\end{itemize}

\subsection{GUI Grounding Benchmark and Performance}
To systematically evaluate GUI grounding under diverse user instructions, we conduct experiments on five benchmarks with different instruction characteristics and visual complexities: ScreenSpot-Pro~\cite{li2025screenspotpro}, ScreenSpot-V2~\cite{osaltas_and_screenspot_v2}, MMBench-GUI L2~\cite{mmbenchgui}, UI-Vision~\cite{ui_vision}, and OSWorld-G-R~\cite{jedi}, which cover desktop, mobile, and web interfaces from complementary perspectives. As shown in Table~\ref{tab:main_grounding}, BiGrounder-4B achieves competitive results across all five benchmarks, reaching 66.6\% on the high-resolution ScreenSpot-Pro and 87.5\% on MMBench-GUI L2. BiGrounder is particularly robust on benchmarks whose instructions carry complex or implicit user intent, consistent with its more stable grounding under instruction variation.

% TODO: reinsert Table (ScreenSpot-Pro: tables/sspro_r.tex, \label{tab:screenspot_pro}) --- convert to two-column (table*) when reinserting.
% TODO: reinsert Table (MMBench-GUI L2: tables/mmbenchGUI.tex) --- convert to two-column (table*) when reinserting.
% TODO: reinsert Table (UI-Vision: tables/UI-vision_r.tex) --- convert layout when reinserting.
% TODO: reinsert Table (OSWorld-G-refined: tables/osworld-g-refined.tex, \label{tab:osworld_g_refine}) --- convert layout when reinserting.

\subsection{GUI Agent Performance}
Beyond static grounding, we test whether it transfers to online agent tasks, where the executor must localize targets from instructions dynamically generated by a planner. All models share the same GPT-4o planner and interaction budget, and differences reflect the executor alone. In terms of accuracy, BiGrounder-4B reaches the highest success rate on both benchmarks, indicating that its grounding stays reliable when queries are produced by a planner. In terms of robustness, its run-to-run std is comparable or lower ($0.5$ vs.\ $0.8$ on OSWorld), indicating stable behavior across the varied phrasings a planner produces. The gains are modest but consistent across both metrics.

\input{tables/android_world}

\subsection{Robustness to Instruction Variation}

\input{tables/osworld_g_r_diverse}
To quantify robustness to instruction reformulation, we build \textbf{OSWorld-G-R-Variants} ($564$ elements $\times\,4$ instructions $=2{,}256$ grounding queries), pairing each element's original instruction with three variants rewritten by GPT-4o and manually verified to be semantically equivalent and unambiguous, then querying every model with all four. As shown in Table~\ref{tab:osworld_g_r_diverse}, BiGrounder attains the highest original-instruction accuracy and the highest Any- and All-Correct rates, while reducing the Any--All gap from 17.4 to 14.8 points compared with Qwen3-VL-4B. This 2.6-point reduction indicates lower sensitivity to instruction wording and more consistent grounding across semantically equivalent expressions. We will release these rewritten instructions together with the exact prompts and filtering procedure to support full reproducibility.

\begin{figure}[h]
    \centering
    \begin{subfigure}[t]{0.49\linewidth}
        \centering
        \includegraphics[width=\linewidth]{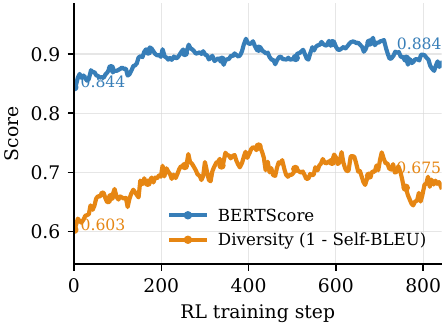}
        \caption{Semantic consistency and within-group diversity.}
        \label{fig:rl_dynamics_a}
    \end{subfigure}
    \hfill
    \begin{subfigure}[t]{0.49\linewidth}
        \centering
        \includegraphics[width=\linewidth]{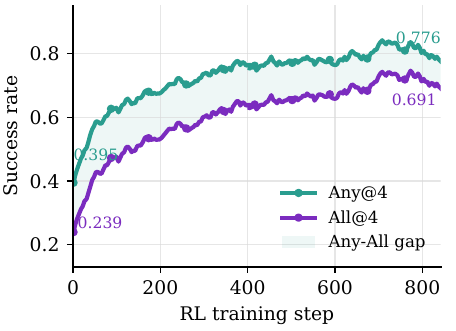}
        \caption{Cycle-grounding reliability.}
        \label{fig:rl_dynamics_b}
    \end{subfigure}
    \caption{RL training dynamics. \textbf{(a)} semantic consistency (BERTScore) and within-group diversity ($1-$Self-BLEU) of the on-policy instructions. \textbf{(b)} cycle-grounding reliability measured by Any@4 and All@4, with the shaded Any--All gap narrowing over training.}
    \label{fig:rl_dynamics}
\end{figure}

\paragraph{\textbf{Training Dynamics. }}
Figure~\ref{fig:rl_dynamics} tracks the on-policy instructions over RL. In Fig.~\ref{fig:rl_dynamics_a}, semantic consistency and diversity rise together (BERTScore $0.844\!\to\!0.884$, diversity $0.603\!\to\!0.675$), suggesting the model explores more varied phrasings while staying close to the target. In Fig.~\ref{fig:rl_dynamics_b}, both Any@4 and All@4 improve, where Any@4 and All@4 are the accuracies of grounding an element correctly under at least one and under all four instructions; the shaded gap between them narrows from about $0.156$ to $0.085$, indicating grounding becomes less dependent on the specific wording.

\begin{figure}[h]
    \centering
    \includegraphics[width=\columnwidth, trim=0 0 0 0, clip]{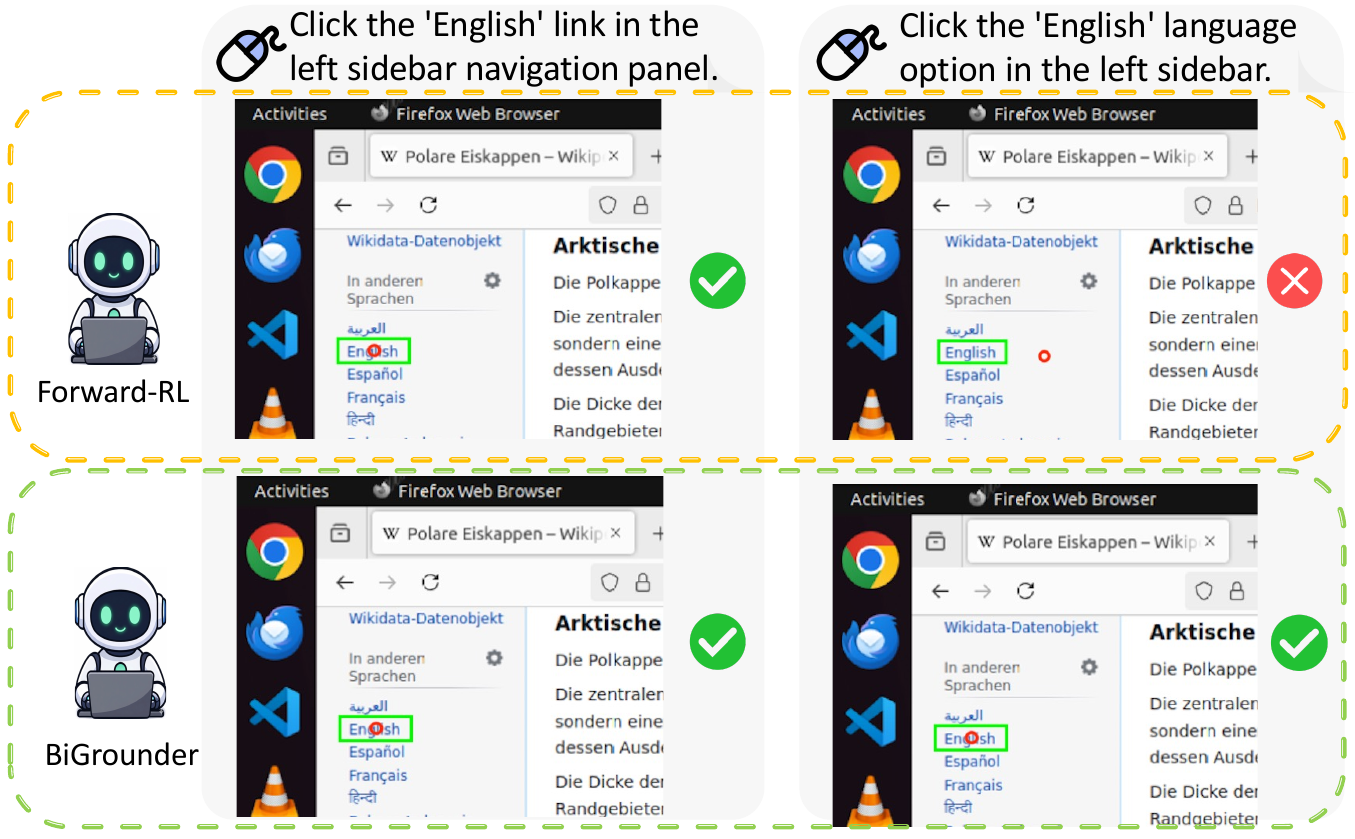}
    \caption{Qualitative comparison of grounding under instruction variations.}
    \label{fig:badcase-forward}
\end{figure}

\paragraph{\textbf{Qualitative Analysis. }}
Figure~\ref{fig:badcase-forward} provides an illustrative example in which forward-only RL is affected by instruction reformulation, whereas BiGrounder consistently grounds the target across the two semantically equivalent expressions.

% TODO: reinsert Figure (semantic generation before/after RL: figures/badcase.pdf, \label{fig:badcase-reverse}) --- redraw for two-column layout.

% TODO: reinsert paragraph below together with figures/badcase.pdf (\label{fig:badcase-reverse}) when the figure is redrawn.
% As illustrated in Fig.~\ref{fig:badcase-reverse}, before RL, the model exhibits spatial ambiguity in dense UI regions, frequently misattributing target coordinates to adjacent elements such as "Activities". Following RL optimization, the model effectively mitigates this semantic drift. The generated instructions achieve precise visual-text grounding on the correct target---specifically the "Slack" icon---while successfully leveraging diverse spatial references, including relative positioning to the "Slack" icon or "Activities" text. This highlights the method's robustness in generating accurate and syntactically varied localized instructions.

%% file: tables/grounding_summary_backbone.tex
% ============================================================
% Main grounding table grouped by BACKBONE (Qwen2.5-VL-3B / Qwen2.5-VL-7B / Qwen3-VL).
% Baselines from MAI-UI report (arXiv:2512.22047) and GroundNext; "$*$" = re-evaluated by us.
% Best in bold, second-best underlined, WITHIN each backbone group.
% ============================================================
\providecommand{\h}[1]{\textbf{\footnotesize \textsf{#1}}}
\providecolor{gainGreen}{HTML}{1A9850}
\providecommand{\gain}[1]{\textcolor{gainGreen}{$_{+#1}$}}

\begin{table*}[!t]
    \centering
    \small
    \setlength{\tabcolsep}{6pt}
    \renewcommand{\arraystretch}{1.12}
    \resizebox{\textwidth}{!}{%
    \begin{tabular}{lcccccc}
        \toprule
        \h{Model} & \h{ScreenSpot-Pro} & \h{ScreenSpot-V2} & \h{MMBench-GUI L2} & \h{UI-Vision} & \h{OSWorld-G-R} & \h{Avg.} \\
        \midrule
        \rowcolor{gray!15}
        \multicolumn{7}{l}{\textit{Qwen2.5-VL-3B backbone}} \\
        Qwen2.5-VL-3B$^*$~\cite{qwen25vl}                      & 25.9 & 80.8 & 52.0 & 0.7  & 39.4 & 39.8 \\
        UI-R1-E-3B~\cite{ui_r1}                                & 17.8 & 88.6 & 68.4 & 16.5 & 48.8 & 48.0 \\
        SE-GUI-3B~\cite{SE_GUI}                                & 35.9 & 86.8 & 66.3 & 15.0 & 46.1 & 50.0 \\
        InfiGUI-R1-3B~\cite{liu2025infiguig1}                  & 35.7 & 89.5 & 70.6 & 17.8 & 42.9 & 51.3 \\
        GUI-G$^2$-3B~\cite{GUI_G2}                             & 36.4 & 87.6 & 66.3 & 18.7 & \underline{53.5} & 52.5 \\
        GUI-G1-3B~\cite{GUIG1}                                 & 37.1 & 89.5 & 71.0 & 20.3 & 49.5 & 53.5 \\
        InfiGUI-G1-3B~\cite{liu2025infiguig1}                  & \underline{45.2} & \underline{91.1} & \underline{73.4} & \underline{22.0} & 49.6 & \underline{56.3} \\
        BiGrounder-3B (Ours)                                   & \textbf{52.5} & \textbf{91.4} & \textbf{78.7} & \textbf{28.7} & \textbf{58.6} & \textbf{62.0}\gain{22.2} \\
        \midrule
        \rowcolor{gray!15}
        \multicolumn{7}{l}{\textit{Qwen2.5-VL-7B backbone}} \\
        Qwen2.5-VL-7B$^*$~\cite{qwen25vl}                      & 27.6 & 86.5 & 69.9 & 0.9  & 41.3 & 45.2 \\
        SE-GUI-7B~\cite{SE_GUI}                                & 47.3 & 68.9 & 34.5 & 16.7 & 33.9 & 40.3 \\
        UI-TARS-1.5-7B~\cite{ui-tars-15}                       & 49.6 & 90.3 & 64.3 & 20.8 & \underline{64.2} & 57.8 \\
        GUI-G$^2$-7B~\cite{GUI_G2}                             & 47.5 & \underline{93.3} & 79.5 & 25.6 & 61.9 & 61.6 \\
        InfiGUI-G1-7B~\cite{liu2025infiguig1}                  & \underline{51.9} & \textbf{93.5} & \underline{80.8} & \underline{26.1} & 59.9 & 62.4 \\
        GTA1-7B~\cite{GTA1}                                    & 50.1 & 92.4 & 79.4 & 25.7 & \textbf{67.7} & \underline{63.1} \\
        BiGrounder-7B (Ours)                                   & \textbf{53.8} & 92.7 & \textbf{81.4} & \textbf{32.1} & 60.7 & \textbf{64.1}\gain{18.9} \\
        \midrule
        \rowcolor{gray!15}
        \multicolumn{7}{l}{\textit{Qwen3-VL-4B backbone}} \\
        % Qwen3-VL-2B$^*$~\cite{Qwen3-VL}                        & 48.5 & 88.5 & 46.5 & 6.2  & 57.4 & 49.4 \\
        % MAI-UI-2B~\cite{zhou2025maiuitechnicalreportrealworld} & 57.4 & 92.5 & 82.6 & 30.3 & 63.5 & 65.3 \\
        Qwen3-VL-4B$^*$~\cite{Qwen3-VL}                        & 59.3 & 93.5 & 81.2 & 26.9 & 65.3 & 65.2 \\
        % MAI-UI-8B~\cite{zhou2025maiuitechnicalreportrealworld} & \underline{65.8} & \textbf{95.2} & \textbf{88.8} & \underline{40.7} & \underline{68.6} & \underline{71.8} \\
        Step-GUI-4B~\cite{step-gui}                            & 60.0 & \underline{93.6} & \underline{84.0} & $27.1^*$ & 67.6$^*$ & 66.5 \\
        GUI-Owl-1.5-4B-Instruct~\cite{gui-owl-1.5}                 & \textbf{66.8} & 93.2 & 83.2 & \underline{$27.3^*$} & \underline{68.4} & \underline{67.8} \\
        BiGrounder-4B (Ours)                                   & \underline{66.6} & \textbf{93.9} & \textbf{87.5} & \textbf{46.6} & \textbf{69.5} & \textbf{72.8}\gain{7.6} \\
        \bottomrule
    \end{tabular}%
    }
    \caption{Grounding accuracy on five GUI grounding benchmarks and their average, grouped by backbone. Unless marked with ``$*$'', baseline results are taken from the corresponding original papers or technical reports (cited in each row), or from the consolidated evaluation reported by GroundNext~\cite{feizi2025groundingcomputeruseagents}. ``$*$'' marks results we re-evaluated under our unified protocol for benchmarks not reported in these sources. ``Avg.'' is the mean over the five benchmarks. Within each backbone group, \textbf{best} is in bold and \underline{second-best} is underlined. The green subscript denotes the absolute Avg.\ gain over each model's backbone.}
    \vspace{-0.2cm}
    \label{tab:main_grounding}
\end{table*}

%% file: tables/android_world.tex
% ============================================================
% Main-experiment table: GUI agent (planner+executor) results.
% Simple 3-column layout: Executor Model | AndroidWorld | OSWorld.
% Baselines from the agentic-workflow section (GPT-4o planner):
%   AndroidWorld = predefined-steps column; OSWorld = 50-step column.
% Self-contained colors (\providecolor); AAAI-safe (no negative \vspace).
%       best in bold.
% ============================================================
\providecolor{ratioTwenty}{HTML}{F7FBFE}
\providecolor{ratioForty}{HTML}{EDF4FA}

\begin{table}[h]
\centering
\footnotesize
\setlength{\tabcolsep}{6pt}
\renewcommand{\arraystretch}{1.15}
\resizebox{\columnwidth}{!}{%
\begin{tabular}{lcc}
\toprule
\textbf{Executor Model} & \textbf{AndroidWorld} & \textbf{OSWorld} \\
\midrule
Aria-UI~\cite{ariaui}                & 44.8 & 15.2 \\
% OS-Atlas-7B~\cite{osaltas_and_screenspot_v2} & 30.1 & 14.6 \\
UGround-V1-7B~\cite{uground}         & 32.8 & 16.1 \\
UI-TARS-1.5-7B~\cite{ui-tars-15}     & 37.9 & 19.1 \\
% JEDI-7B~\cite{jedi}                  & 43.7 & 25.0 \\
ScaleCUA-7B~\cite{liu2025scalecua}   & 48.3 & 28.1 \\
\midrule
\rowcolor{ratioTwenty}
Qwen3-VL-4B$^{*}$~\cite{Qwen3-VL}    & 68.9 $\pm$ 0.9 & 40.2 $\pm$ 0.8 \\
\rowcolor{ratioForty}
BiGrounder-4B  & \textbf{72.4 $\pm$ 0.9} & \textbf{42.8 $\pm$ 0.5} \\
\bottomrule
\end{tabular}%
}
\caption{GUI agent benchmark results with each model as the grounding executor and GPT-4o as the planner. AndroidWorld follows predefined steps; OSWorld uses a 50-step budget. \textbf{Best} in bold. $^{*}$Evaluated by us (mean $\pm$ std over 3 runs).}
\vspace{-0.3cm}
\label{tab:gui_agent_benchmark}
\end{table}

%% file: tables/osworld_g_r_diverse.tex
% ============================================================
% Main-experiment table: paraphrase-consistency on OSWorld-G-R-Variants.
% Purpose: quantify robustness to instruction variation across four instructions
% for the SAME element -- the original dataset instruction plus three rewritten
% by an external LLM from three perspectives (user intent / icon style /
% spatial position).
% Backbone: Qwen3-VL-4B. Qwen3-VL-4B baseline vs BiGrounder-4B.
% Self-contained colors (\providecolor); AAAI-safe (no negative \vspace).
%   Original Accuracy = grounding accuracy using the original dataset instruction (%)          higher better
%   Any-Correct Rate  = fraction of elements grounded correctly on AT LEAST ONE of the 4 (%)   higher better  (upper bound)
%   All-Correct Rate  = fraction of elements grounded correctly on ALL 4 (%)                   higher better  (lower bound / worst case)
%   Any--All Gap      = Any-Correct minus All-Correct (percentage points)                       lower = more robust
% ============================================================
\providecolor{ratioTwenty}{HTML}{F7FBFE}
\providecolor{ratioThirty}{HTML}{F2F7FB}
\providecolor{ratioForty}{HTML}{EDF4FA}
\providecolor{ratioFifty}{HTML}{E7F0F8}
\providecommand{\gain}[1]{\textcolor{gainGreen}{$_{+#1}$}}
\providecommand{\gapreduce}[1]{\textcolor{gainGreen}{$_{\downarrow #1}$}}
\providecolor{gainGreen}{HTML}{1A9850}

\begin{table}[h]
\centering
\footnotesize
\setlength{\tabcolsep}{1.1pt}
\renewcommand{\arraystretch}{1.2}
\begin{tabular}{lcccc}
\toprule
\textbf{Method} & \makecell[c]{\textbf{Original}\\\textbf{Accuracy}~$\uparrow$} & \makecell[c]{\textbf{Any}\\\textbf{Correct}~$\uparrow$} & \makecell[c]{\textbf{All}\\\textbf{Correct}~$\uparrow$} & \makecell[c]{\textbf{Any--All}\\\textbf{Gap}~$\downarrow$} \\
\midrule
\rowcolor{ratioTwenty}
Qwen3-VL-4B             & 65.3 & 73.4 & 56.0 & 17.4 \\
\rowcolor{ratioThirty}
Bi-SFT                   & 65.7 & 73.6 & 56.8 & 16.8 \\
\rowcolor{ratioForty}
Bi-SFT + Forward RL      & 67.4 & 74.5 & 58.2 & 16.3 \\
\rowcolor{ratioFifty}
BiGrounder-4B             & \textbf{69.5} & \textbf{75.3} & \textbf{60.5} & \textbf{14.8}\gapreduce{2.6} \\
\bottomrule
\end{tabular}
\caption{Robustness to instruction variation on \textbf{OSWorld-G-R-Variants}. Each original instruction is paired with three target-preserving variants that differ in wording and sentence structure. \textbf{Original}: accuracy under the original instruction; \textbf{Any-/All-Correct}: element grounded correctly under at least one/all four instructions. \textbf{Any--All Gap}: sensitivity to instruction variation. The green subscript denotes the absolute gap reduction over Qwen3-VL-4B. \textbf{Best} in bold.}
\label{tab:osworld_g_r_diverse}
\end{table}

%% file: sections/ablation.tex
\section{Ablation Studies}

\input{tables/al_all_data}

\paragraph{\textbf{Effectiveness of Training Recipes.}}
Table~\ref{tab:qwen3vl_ablation} shows that Bi-SFT improves over Single-SFT ($67.6$ vs.\ $66.9$). Adding bidirectional RL (\textbf{+RL-Bi}) yields the best average of $72.8$, outperforming three point-reward alternatives: a binary in-box hit (\textbf{InBbox}), a distance-to-center decay (\textbf{Distance}), and a Gaussian distance reward (\textbf{Gaussian}). To isolate the contribution of the reverse branch, we further evaluate \textbf{+RL-Forward-only}, a forward-only variant that uses the same reward without reverse policy updates.

\input{tables/BertScore}
\paragraph{\textbf{Effect of Cycle-Consistency Gating.}}
BERTScore alone rewards semantically fluent instructions even when they are visually incorrect, leaving the reverse reward open to reward hacking. To counter this, Algorithm~\ref{alg:cycle_rl} retains the reverse reward only when the generated instruction can be grounded back to the target element, so that a high semantic score alone cannot earn credit unless the instruction is also verified to point back to the intended element. As reported in Table~\ref{tab:reward_gating}, cycle gating raises the average from $70.1$ to $72.8$ over the ungated variant, confirming that filtering out non-groundable instructions yields more reliable reverse rewards.

% \input{tables/generation_quality}

\input{tables/wo_gemini}
\paragraph{\textbf{Effect of Gemini-Rewritten Instructions.}}
Table~\ref{tab:gemini_ablation} isolates the Gemini-rewritten instructions under the same backbone and recipe. Removing them lowers the average only marginally at both stages ($67.6\!\to\!67.3$ SFT, $72.8\!\to\!71.3$ RL), indicating the gains stem from bidirectional training rather than the external rewrites.

\begin{figure}[!t]
    \centering
    \begin{subfigure}[t]{0.49\linewidth}
        \centering
        \includegraphics[height=2.68cm]{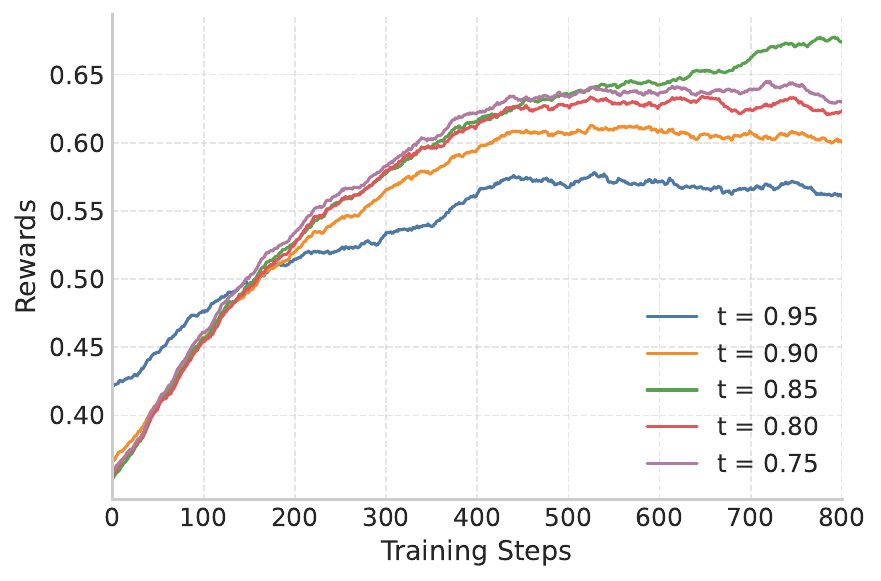}
        \caption{Mean reward during training.}
        \label{fig:al_t_reward}
    \end{subfigure}
    \hfill
    \begin{subfigure}[t]{0.49\linewidth}
        \centering
        \includegraphics[height=2.68cm]{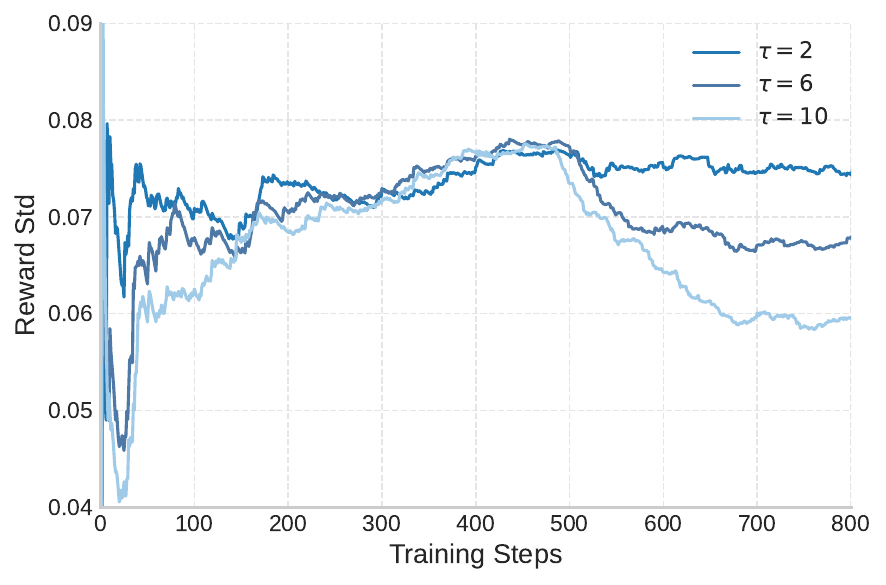}
        \caption{Reward std. during training.}
        \label{fig:al_tau}
    \end{subfigure}
    \caption{Training dynamics under different reward hyperparameters: (a) semantic-reward threshold $t$; (b) pixel-level tolerance $\tau$.}
    \label{fig:al_t}
\end{figure}

\paragraph{\textbf{Sensitivity to Reward Hyperparameters.}}
Using a ScaleCUA subset and a fixed configuration, $t=0.85$ gives the highest stabilized reward (Figure~\ref{fig:al_t}(a)). A strict tolerance ($\tau=2$) causes excessive variation, whereas $\tau=10$ weakens discrimination (Figure~\ref{fig:al_t}(b)); we therefore use $\tau=6$.

%% file: tables/al_all_data.tex
% ============================================================
% Ablation: controlled comparison of training recipes (Qwen3-VL-4B backbone).
% Source: ECCV rebuttal tables/al_all_data.tex
% Self-contained: defines its own row-shading colors (\providecolor) so it needs
% no extra preamble beyond colortbl (\rowcolor), already loaded in main.tex.
% AAAI-safe: no negative \vspace near the table/caption.
% ============================================================
\providecolor{ratioTwenty}{HTML}{F7FBFE}
\providecolor{ratioThirty}{HTML}{F2F7FB}
\providecolor{ratioForty}{HTML}{EDF4FA}
\providecolor{ratioFifty}{HTML}{E7F0F8}
\providecolor{ratioSeventy}{HTML}{D6E6F3}
\providecolor{gainGreen}{HTML}{1A9850}
\providecommand{\gain}[1]{\textcolor{gainGreen}{$_{+#1}$}}

\begin{table}[!t]
\centering
\footnotesize
\setlength{\tabcolsep}{1.2pt}
\resizebox{\columnwidth}{!}{%
\begin{tabular}{lcccccc}
\toprule
\textbf{Method} & \textbf{SS-Pro} & \textbf{SS-V2} & \textbf{OS-G-R} & \textbf{UI-V.} & \textbf{MM-L2} & \textbf{Avg.} \\
\midrule
\rowcolor{ratioTwenty}
Qwen3-VL-4B    & 59.3 & 93.5 & 65.3 & 26.9 & 81.2 & 65.2 \\
\rowcolor{ratioThirty}
Single-SFT     & 59.4 & 90.6 & 65.2 & 39.7 & 79.8 & 66.9\gain{1.7} \\
\rowcolor{ratioThirty}
Bi-SFT & 60.2 & 91.2 & 65.7 & 39.9 & 81.0 & 67.6\gain{2.4} \\
\rowcolor{ratioForty}
\quad +RL-InBbox     & 63.2 & 92.9 & 67.4 & 44.9 & 84.1 & 70.5\gain{5.3} \\
\rowcolor{ratioForty}
\quad +RL-Distance   & 62.7 & 92.3 & 66.8 & 42.5 & 83.5 & 69.6\gain{4.4} \\
\rowcolor{ratioForty}
\quad +RL-Gaussian   & 63.5 & 92.9 & 67.9 & 44.8 & 85.1 & 70.8\gain{5.6} \\
\rowcolor{ratioFifty}
\quad +RL-Forward-only   & 63.7 & 92.9 & 67.4 & 45.1 & 85.3 & 70.9\gain{5.7} \\
\rowcolor{ratioFifty}
\quad +RL-Bi (Ours)   & 66.6 & 93.9 & 69.5 & 46.6 & 87.5 & \textbf{72.8}\gain{7.6} \\
\bottomrule
\end{tabular}%
}
\caption{Controlled comparison of training recipes with Qwen3-VL-4B-Instruct backbone and complete training dataset. The subscript in Avg. denotes the absolute gain over the base model.}
\label{tab:qwen3vl_ablation}
\end{table}

%% file: tables/BertScore.tex
% ============================================================
% Ablation: cycle-consistency gating for the reverse reward.
% Source: ECCV rebuttal tables/BertScore.tex
% Self-contained colors (\providecolor); AAAI-safe (no negative \vspace).
% NOTE: the "RL w/ Semantic Reward" row values are placeholders ("-") to fill in.
% ============================================================
\providecolor{ratioTwenty}{HTML}{F7FBFE}
\providecolor{ratioThirty}{HTML}{F2F7FB}
\providecolor{ratioForty}{HTML}{EDF4FA}
\providecolor{gainGreen}{HTML}{1A9850}
\providecommand{\gain}[1]{\textcolor{gainGreen}{$_{+#1}$}}

\begin{table}[h]
\centering
\footnotesize
\setlength{\tabcolsep}{2.2pt}
\resizebox{\columnwidth}{!}{%
\begin{tabular}{lcccccc}
\toprule
\textbf{Method} & \textbf{SS-Pro} & \textbf{SS-V2} & \textbf{OS-G-R} & \textbf{UI-V.} & \textbf{MM-L2} & \textbf{Avg.} \\
\midrule
\rowcolor{ratioTwenty}
Bi-SFT & 60.2 & 91.2 & 65.7 & 39.9 & 81.0 & 67.6 \\
\rowcolor{ratioThirty}
RL w/o Cycle-C. & 62.8 & 92.4 & 66.9 & 44.9 & 83.5 & 70.1\gain{2.5} \\
% \rowcolor{ratioThirty}
% RL w/ Semantic Reward & - & - & - & - & - & - \\
\rowcolor{ratioForty}
RL w/ Cycle-C.  & 66.6 & 93.9 & 69.5 & 46.6 & 87.5 & \textbf{72.8}\gain{5.2} \\
\bottomrule
\end{tabular}%
}
\caption{Ablation on cycle-consistency gating for mitigating reward hacking. The subscript in Avg. denotes the absolute gain over the Bi-SFT baseline.}
\label{tab:reward_gating}
\end{table}

%% file: tables/wo_gemini.tex
% \begin{table}[h]
% \centering
% \footnotesize
% \setlength{\tabcolsep}{3.0pt}
% \caption{Ablation on Gemini-generated data using Qwen3-VL-4B-Instruct. We report results on five benchmarks and the average score.}
% \label{tab:gemini_ablation}
% \begin{tabular}{lcccccc}
% \toprule
% \textbf{Method} & \textbf{SS-Pro} & \textbf{SS-V2} & \textbf{OS-G-R} & \textbf{UI-V.} & \textbf{MM-L2} & \textbf{Avg.} \\
% \midrule
% \rowcolor{ratioTwenty}
% Qwen3-VL-4B       & 59.3 & 94.0 & 65.3 & 26.9 & 81.2 & 65.3 \\
% \rowcolor{ratioThirty}
% w/o Gemini + SFT  & 59.9 & 91.7 & 65.5 & 41.5 & 80.8 & 67.9 \\
% \rowcolor{ratioForty}
% w/o Gemini + RL   & 65.1 & 93.2 & 68.7 & 45.1 & 85.3 & 71.5 \\
% \rowcolor{ratioFifty}
% w/ Gemini + SFT   & 60.2 & 91.7 & 65.7 & 41.2 & 81.0 & 68.0 \\
% \rowcolor{ratioSeventy}
% w/ Gemini + RL    & 66.6 & 93.9 & 69.5 & 46.6 & 87.5 & 72.8 \\
% \bottomrule
% \end{tabular}
% \end{table}
\providecolor{ratioTwenty}{HTML}{F7FBFE}
\providecolor{ratioThirty}{HTML}{F2F7FB}
\providecolor{ratioForty}{HTML}{EDF4FA}
\providecolor{ratioFifty}{HTML}{E7F0F8}
\begin{table}[h]
\centering
\footnotesize
\setlength{\tabcolsep}{1.6pt}
\resizebox{\columnwidth}{!}{%
\begin{tabular}{lcccccc}
\toprule
\textbf{Method} & \textbf{SS-Pro} & \textbf{SS-V2} & \textbf{OS-G-R} & \textbf{UI-V.} & \textbf{MM-L2} & \textbf{Avg.} \\
\midrule
\rowcolor{ratioTwenty}
Qwen3-VL-4B        & 59.3 & 93.5 & 65.3 & 26.9 & 81.2 & 65.2 \\
\midrule
\rowcolor{ratioThirty}
Bi-SFT w/o rewritten     & 59.7 & 91.2 & 65.5 & 39.5 & 80.6 & 67.3 \\
\rowcolor{ratioFifty}
Bi-SFT w/ rewritten      & 60.2 & 91.2 & 65.7 & 39.9 & 81.0 & 67.6 \\
\midrule
\rowcolor{ratioThirty}
BiGrounder w/o rewritten  & 64.7 & 93.1 & 68.1 & 44.8 & 85.6 & 71.3 \\
\rowcolor{ratioFifty}
BiGrounder w/ rewritten   & 66.6 & 93.9 & 69.5 & 46.6 & 87.5 & 72.8 \\
\bottomrule
\end{tabular}%
}
\caption{Effect of Gemini-rewritten instructions under the same backbone and training recipe.}
\label{tab:gemini_ablation}
\end{table}

%% file: sections/conclusion.tex
\section{Conclusion}
We present BiGrounder, a bidirectional training paradigm for the many-to-one instruction mapping in GUI grounding. It turns the model's own on-policy instruction variations into reinforcement trajectories. Across five benchmarks, our model achieves competitive average accuracy in our evaluation, with the largest gains in semantically challenging scenarios, and plugs readily into standard SFT+RL pipelines.

%% file: appendix/appendix.tex
% Appendix body shared by standalone supplementary and full-paper builds.
% Load appendix_preamble.tex in the parent document's preamble, then input
% this file once after the main text. The parent document owns the title,
% author list, bibliography, and document environment.
\appendix
\raggedbottom
\makeatletter
\setlength{\@dblfptop}{0pt}
\setlength{\@dblfpsep}{12pt plus 2pt minus 2pt}
\makeatother
\setcounter{secnumdepth}{2}
\counterwithin{figure}{section}
\counterwithin{table}{section}
\counterwithin{equation}{section}

% Loosen float placement so full-width (table*) tables may sit on a page that
% also carries text, instead of being pushed onto near-empty float-only pages.
\renewcommand{\topfraction}{0.95}
\renewcommand{\bottomfraction}{0.95}
\renewcommand{\textfraction}{0.05}
\renewcommand{\floatpagefraction}{0.85}
\renewcommand{\dbltopfraction}{0.95}
\renewcommand{\dblfloatpagefraction}{0.85}
\setcounter{topnumber}{4}
\setcounter{bottomnumber}{4}
\setcounter{totalnumber}{8}
\setcounter{dbltopnumber}{4}

\section{Qualitative Results on Cycle-Consistent Reinforcement}
\label{app:qualitative}
\begin{figure*}[t]
    \centering
    \includegraphics[width=\textwidth]{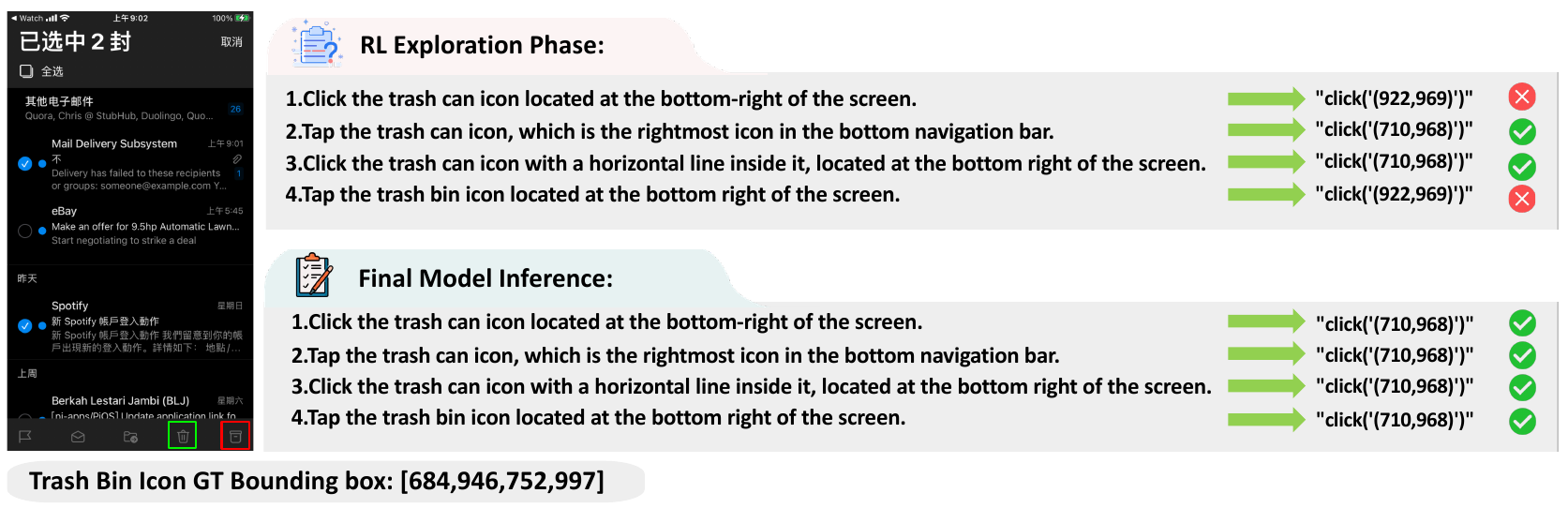}
    \caption{Instruction following comparison before and after cycle-consistency reinforcement, showing improved spatial grounding.}
    \label{fig:0311_1}
\end{figure*}

\begin{figure*}[t]
    \centering
    \includegraphics[width=\textwidth]{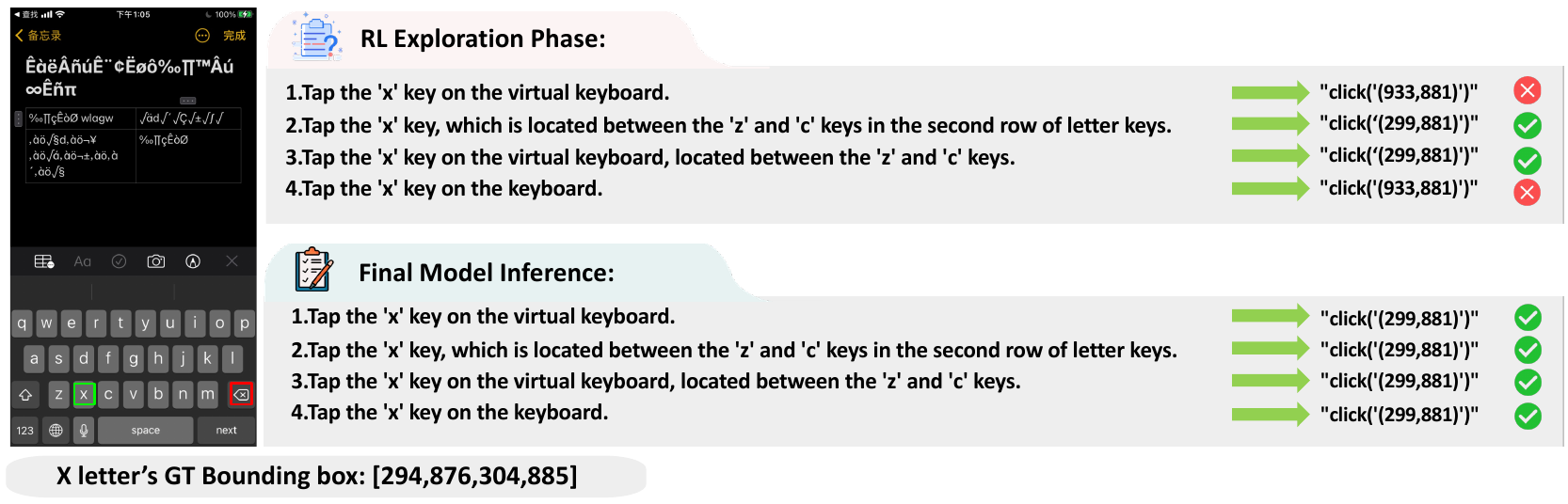}
    \caption{Instruction following comparison before and after cycle-consistency reinforcement, showing improved icon grounding.}
    \label{fig:0311_2}
\end{figure*}

During the reinforcement process, we observe a critical phenomenon:
despite the high semantic similarity among many instructions, the
agent still fails to perform correct grounding. Our framework aims
to improve the robustness of instruction execution. Fig.~\ref{fig:0311_1} and Fig.~\ref{fig:0311_2} present
real cases from the reinforcement process, where the model makes
grounding errors when handling similar instructions. Through our
proposed cycle-consistency reinforcement framework, after multiple
iterations, the model's generalization ability on similar instructions
is significantly improved.

\begin{figure*}[!t]
    \centering
    \includegraphics[width=\textwidth]{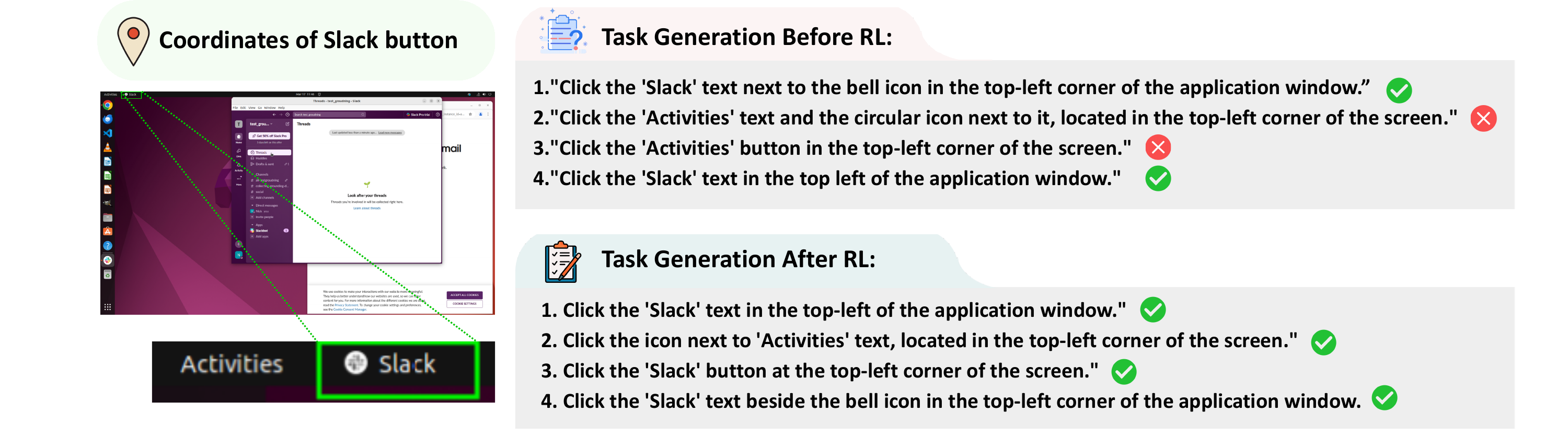}
    \caption{Analyzing the semantic generation capability before and after reinforcement learning.}
    \label{fig:badcase}
\end{figure*}

As illustrated in Fig.~\ref{fig:badcase}, before RL, the model exhibits spatial ambiguity in dense UI regions, frequently misattributing target coordinates to adjacent elements such as ``Activities''. Following RL optimization, the model effectively mitigates this semantic drift. The generated instructions achieve precise visual--text grounding on the correct target---specifically the ``Slack'' icon---while successfully leveraging diverse spatial references, including relative positioning to the ``Slack'' icon or ``Activities'' text. This highlights the method's robustness in generating accurate and syntactically varied localized instructions.

\section{Dataset Construction Details}
\label{app:dataset}

\begin{figure}[!t]
    \centering
    \includegraphics[width=\columnwidth]{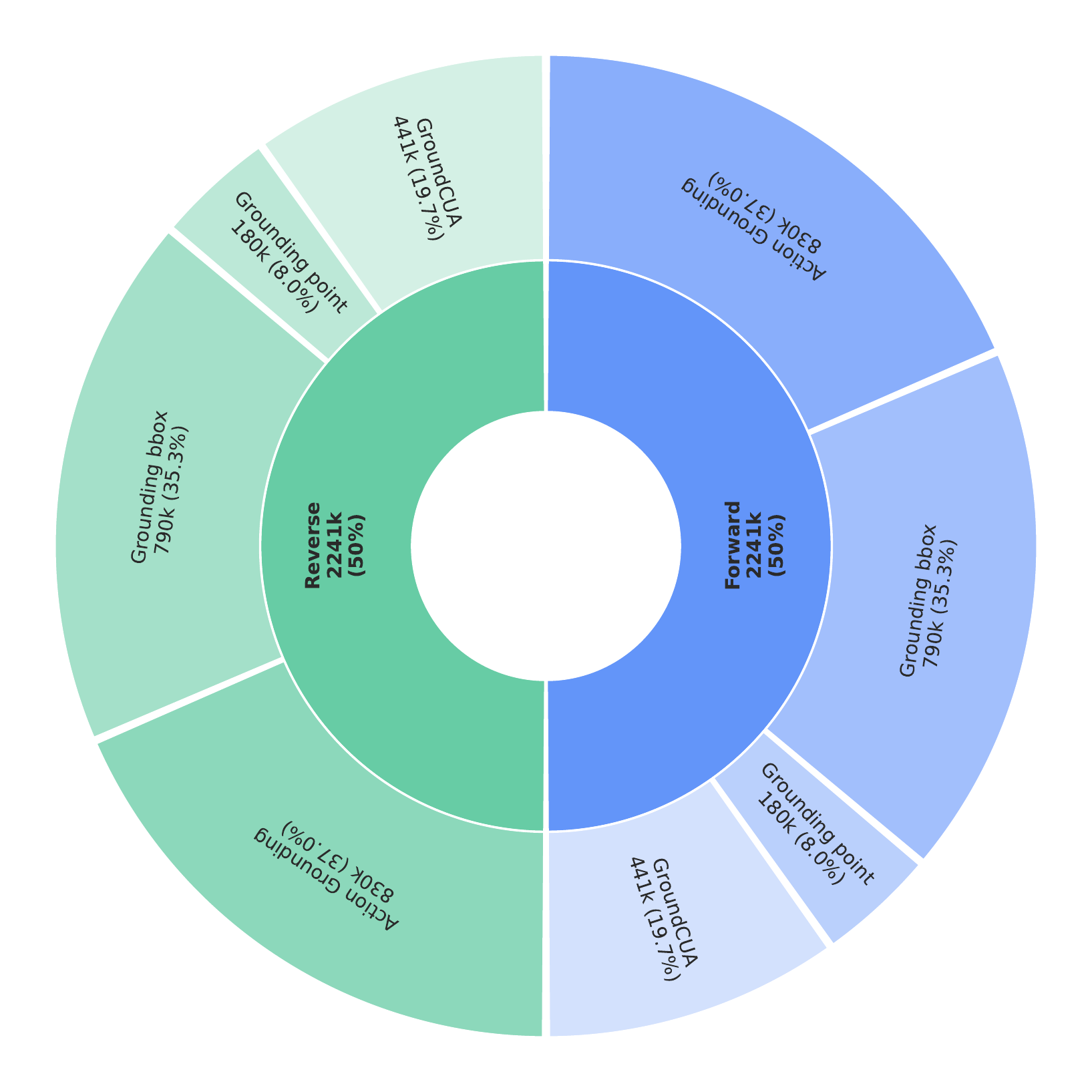}
    \caption{Composition of the 4{,}482k-pair training corpus. The inner ring splits the data into forward (instruction$\rightarrow$coordinate) and reverse (coordinate$\rightarrow$instruction) halves of equal size; the outer ring breaks each half down by data source and grounding modality. The GroundCUA portion uses \emph{Video-Audio Production} samples and their reverse counterparts. As an optional augmentation, the 790k \emph{Grounding bbox} samples---those that already carry bounding-box annotations---have their instructions rewritten with Gemini-2.5-Flash to broaden localization-based reward coverage during RL; the core pipeline can be trained on open-source data alone.}
    \label{fig:data_composition}
\end{figure}

Fig.~\ref{fig:data_composition} summarizes the composition of our training corpus, which comprises \textbf{4{,}482k} instruction--grounding pairs in total. The corpus is split evenly into two symmetric halves that mirror the bidirectional design of BiGrounder: a \emph{forward} half (2{,}241k, 50\%) that maps textual instructions to spatial targets, and a \emph{reverse} half (2{,}241k, 50\%) that generates instructions from ground-truth coordinates. Because every reverse sample is constructed from its corresponding forward grounding sample, the two halves share an identical internal breakdown, which keeps the forward and reverse policies exposed to matched data distributions during training.

Within each half, the data are drawn from two sources. The larger source, ScaleCUA~\cite{liu2025scalecua} (1{,}800k per half), spans the three grounding modalities used throughout our prompts: \emph{Action Grounding} (830k, 37.0\%), \emph{Grounding bbox} (790k, 35.3\%), and \emph{Grounding point} (180k, 8.0\%). The GroundCUA portion consists of \emph{Video-Audio Production} samples and their reverse counterparts: 441k forward action-grounding pairs and 441k reverse instruction-generation pairs (19.7\% of each half). The core pipeline can be trained using open-source data alone. However, since ScaleCUA's action-grounding subset lacks bounding boxes, the coverage of localization-based rewards during RL training is limited. We therefore use Gemini-2.5-Flash \emph{only as an optional augmentation}: it rewrites instructions for the samples that already carry bounding-box annotations, i.e., the 790k \emph{Grounding bbox} portion, using the dual-view refinement pipeline detailed in the ``Grounding Instruction Refine'' subsection below. The remaining modalities retain their original ScaleCUA/GroundCUA formatting, and the generated instructions will be released.

The instruction-generation and grounding prompt templates are collected in Appendix~\ref{app:prompt_templates}.

\subsection{Grounding Instruction Refine}
\label{app:instruction_refinement}
To prepare data for the reinforcement learning stage, we design an automated instruction refinement pipeline utilizing Gemini-2.5-Flash. Since the action-grounding subset of ScaleCUA does not provide bounding-box annotations, the localization-based reward has insufficient action-grounding data to draw on during RL. We therefore repurpose the \emph{Grounding bbox} portion---which already carries bounding-box annotations---by rewriting its instructions into action-oriented commands, thereby filling this gap. We employ a dual-view prompting strategy by feeding the VLM both the global screenshot and a high-resolution crop of the target element. This design enables the model to simultaneously capture global spatial relations derived from the full view and fine-grained visual attributes extracted from the crop. By synthesizing these cues, we transform static region descriptions into precise action-oriented commands suitable for action grounding.
For both bbox and point grounding tasks, we strictly adhere to the data formatting protocols established in ScaleCUA~\cite{liu2025scalecua}.
The refinement prompt is provided in Appendix~\ref{app:prompt_refinement}.

\section{OSWorld-G-Refine-Variants: Construction and Evaluation}
\label{sec:osworld_variants}
To measure how sensitive a grounding model is to the exact wording of an instruction, we augment OSWorld-G-Refine into \textbf{OSWorld-G-Refine-Variants}. This section details how the paraphrase variants are generated and how the paraphrase-consistency metrics reported in the main paper are computed, so that the benchmark can be reproduced exactly.

\subsection{Variant Generation}
Starting from the $564$ annotated elements of OSWorld-G-Refine, we keep each element's original instruction and additionally generate three target-preserving paraphrases, yielding $564\times4=2{,}256$ grounding queries ($564$ original $+\,1{,}692$ variants). Paraphrases are produced by prompting GPT-4o with the template provided in Appendix~\ref{app:prompt_variants}. The template constrains every variant to (i) refer to the \emph{exact same} UI element and location as the original, (ii) use different wording, sentence structure, or perspective, and (iii) preserve the original level of spatial/visual specificity without adding or removing critical cues. To encourage genuine diversity rather than mere synonym substitution, the prompt explicitly lists four rewriting strategies---varying spatial references, emphasizing different visual attributes (color, shape, icon, text, position), changing the interaction verb, and reordering descriptive clauses. The model is queried at a sampling temperature of $0.7$ and must return exactly three paraphrases as a JSON array; malformed responses are re-requested for up to five attempts, and any element whose rewriting still fails falls back to its original instruction and is logged. All generated variants are then manually verified to be semantically equivalent to the source instruction and to remain unambiguous with respect to the target element; ambiguous or drifted paraphrases are corrected before evaluation.

\subsection{Evaluation Protocol}
Each of the $2{,}256$ queries is evaluated independently and then grouped by its source element, so that every group contains the four instructions (one original plus three variants) that point to the same target. Predictions use the same GUI-agent action prompt as our grounding evaluation, decoded greedily (temperature $0$). A predicted coordinate is parsed from the emitted \seqsplit{click(start\_box='(x,y)')} action, mapped from the normalized $[0,1000]$ range back to absolute pixels, and reduced to its center point. A query is counted as correct when this center point lies inside the target region: point-in-rectangle for \texttt{bbox} annotations, point-in-polygon for \texttt{polygon} annotations, and a negative sentinel coordinate for \texttt{refusal} cases (where the correct behavior is to abstain).

On top of the per-query correctness $r$, we report four group-level quantities that together characterize robustness to instruction variation:
\begin{itemize}
    \item \textbf{Original Accuracy}: accuracy computed over the $564$ original instructions only, matching the standard OSWorld-G-Refine setting.
    \item \textbf{Any-Correct Rate}: the fraction of groups in which \emph{at least one} of the four instructions is grounded correctly. This is an upper bound on achievable accuracy and equals Pass@4 over the paraphrase set.
    \item \textbf{All-Correct Rate}: the fraction of groups in which \emph{all four} instructions are grounded correctly---a strict, worst-case measure of consistency.
    \item \textbf{Any--All Gap}: the difference between the Any- and All-Correct rates (in percentage points). A smaller gap indicates that the model's success is less dependent on the particular phrasing of the instruction, i.e., more robust grounding.
\end{itemize}
As an auxiliary diagnostic, we also track the mean intra-group variance $\frac{1}{N}\sum_i p_i(1-p_i)$, where $p_i$ is the fraction of the four instructions in group $i$ that are grounded correctly; this attains $0$ when every group is answered all-right or all-wrong and is maximized ($0.25$) when the model is split evenly within groups. As reported in the main paper, BiGrounder attains the highest Original, Any-Correct, and All-Correct rates and shrinks the Any--All gap from $17.4$ to $14.8$ points relative to its Qwen3-VL-4B backbone, confirming that bidirectional training reduces sensitivity to instruction wording. We will release the rewritten instructions, the exact generation prompt, and the filtering procedure to support full reproducibility.

\section{Reward Hacking and Semantic Noise}
\label{app:reward_hacking}
A central design question in BiGrounder is how to reward the reverse policy $\pi_\theta(\hat{x}\mid s,o^*)$, which generates instructions from a ground-truth target coordinate $o^*$. The natural choice is a semantic similarity reward such as BERTScore between the generated instruction $\hat{x}_i$ and the reference instruction $x$. However, semantic similarity alone is vulnerable to \emph{reward hacking}: the reverse policy can learn to emit instructions that are lexically and semantically fluent, and therefore score high under BERTScore, yet do not actually point to the target element on the screen. Such instructions inject \emph{semantic noise} into the reinforcement loop, because they are treated as valid supervision for the forward grounding policy even though grounding them would lead to the wrong location. If left unchecked, this noise not only wastes the on-policy instruction budget but can actively degrade grounding, since the model is reinforced toward describing the wrong element convincingly rather than the right element accurately.

\paragraph{Why BERTScore alone is insufficient.}
BERTScore measures token-level embedding similarity between $\hat{x}_i$ and $x$ (Eqs.~2--4 in the main paper). We compute BERTScore using RoBERTa-large representations from layer 17. Two failure modes make it an unreliable standalone reward for GUI grounding. First, GUI screens contain many visually and functionally similar elements (e.g., repeated ``star'', ``bookmark'', ``follow'' affordances), so an instruction that describes a plausible-but-wrong element can be nearly indistinguishable from the reference in embedding space. Second, BERTScore rewards surface fluency and paraphrasing, which the policy can exploit to maximize reward without improving visual disambiguation. The case study in Table~\ref{tab:case_reward_hacking} makes this concrete: the distractor ``Click the star to follow the repository.'' attains a high BERTScore of $0.86$ against the ground truth ``Click star to save this page.'', yet it refers to a different action on a different element. A reward that trusts BERTScore in isolation would happily reinforce this instruction.

\paragraph{Cycle-consistency as a grounding-verified gate.}
BiGrounder resolves this by never trusting the semantic score on its own. As formalized in the reverse reward (Eq.~5 of the main paper), the BERTScore of a generated instruction is retained \emph{only if} the instruction survives two gates simultaneously:
\begin{equation*}
R_{\mathrm{rev}}=
\begin{cases}
\operatorname{BERTScore}(\hat{x}_i,x), & b_i\ge t \ \wedge\ c_i=1,\\
0, & \text{otherwise},
\end{cases}
\end{equation*}
where $b_i=F_{\mathrm{BERT}}(\hat{x}_i,x)$ is the semantic score, $t=0.85$ is the semantic threshold, and $c_i=\mathbb{I}[\exists k:\operatorname{success}(\tau_{i,k},o^*)]$ is the \emph{cycle gate}. The cycle gate is the key anti-hacking mechanism: for each candidate instruction $\hat{x}_i$ that passes the semantic threshold, the \emph{same} policy is asked to ground it back with $K$ rollouts $\{\tau_{i,k}\}_{k=1}^{K}$, and $c_i=1$ only if at least one rollout returns to the true target (forward grounding reward $\ge 0.9$). Thus a fluent instruction that cannot be grounded back to $o^*$ receives $R_{\mathrm{rev}}=0$ and is prevented from reinforcing the reverse policy, no matter how high its BERTScore. In the running example, ``Click the star to follow the repository.'' fails the cycle gate ($c_i=0$) and its reward collapses to $0.00$, while the correct paraphrase ``Click the star to bookmark this webpage.'' grounds back successfully ($c_i=1$) and keeps its reward of $0.92$ (Table~\ref{tab:case_reward_hacking}).

\paragraph{Ground-truth fallback keeps the forward loop clean.}
Rejecting a semantically noisy instruction does not waste the sample. When a candidate fails the semantic threshold ($b_i<t$), BiGrounder substitutes the ground-truth instruction $x$ into that forward loop ($q_i\leftarrow x$, the GT fallback in Algorithm~1). Consequently, every one of the $G$ instruction branches still contributes valid forward grounding trajectories, but only semantically qualified \emph{and} cycle-consistent generations are allowed to update the reverse policy. This decoupling is what lets BiGrounder harvest instruction diversity aggressively while keeping the forward supervision signal noise-free.

\paragraph{Quantifying noise removal.}
To verify that cycle-consistency actually suppresses semantic noise rather than merely filtering paraphrases, we manually annotate 300 reverse instructions (273 correct, 27 incorrect) and measure how well each gating strategy separates them. As shown in Table~\ref{tab:P_N}, thresholding on BERTScore alone filters only $18.5\%$ of the incorrect instructions ($5/27$), and even an LLM judge removes only $74.1\%$ ($20/27$). In contrast, the cycle-consistency gate filters $25$ of $27$ incorrect instructions, reaching $92.6\%$ Noise Filtering Rate (NFR) while simultaneously raising Precision to $99.2\%$ on the retained set. This confirms that verifying groundability, rather than semantic plausibility, is the decisive signal for rejecting reward-hacking instructions, and that the two gates are complementary: the semantic threshold discards obviously off-topic generations cheaply, while the cycle gate catches the fluent-but-wrong cases that BERTScore cannot.

\input{appendix/tables/P_N}
\input{appendix/tables/case}

\input{appendix/sections/additional_experiments}

\input{appendix/sections/method_comparison}

\section{Per-Benchmark Detailed Results}
\label{sec:detailed_results}

To complement the aggregated grounding accuracy reported in the main paper, this section provides the full per-category breakdown for each of the five GUI grounding benchmarks. Unless otherwise noted, all BiGrounder-4B results are obtained under the same unified evaluation protocol described in the main paper. Baseline numbers are taken from the corresponding original papers or technical reports, and entries marked with ``$*$'' are re-evaluated by us. In every table, the \textbf{best} result is shown in bold and the \underline{second-best} is underlined. The ``Avg.'' column reports micro accuracy over all benchmark examples, equivalently weighting each split by its number of examples; it therefore need not equal the unweighted arithmetic mean of the displayed split accuracies.

\subsection{ScreenSpot-V2}
Table~\ref{tab:ssv2} reports the full breakdown on ScreenSpot-V2 across the Mobile, Desktop, and Web platforms, each further split into text and icon/widget targets. BiGrounder-4B attains the highest overall average (93.9) and is particularly strong on the Desktop text/icon and Web icon splits, indicating that bidirectional training improves grounding of visually dense, non-textual elements without sacrificing performance on text targets.

\subsection{ScreenSpot-Pro}
ScreenSpot-Pro targets high-resolution professional software with small, densely packed elements. Table~\ref{tab:screenspot_pro} details performance across six application domains (CAD, Development, Creative, Scientific, Office, and OS), each divided into text and icon targets. BiGrounder-4B reaches an average of 66.6, ranking second among all compared models and first among open-source models, with especially large margins on icon targets (e.g., CAD, Development, and OS icons), which are the hardest cases for coordinate prediction.

\subsection{MMBench-GUI L2}
Table~\ref{tab:main_results_mmbench} presents the per-platform results on MMBench-GUI L2, covering Windows, MacOS, Linux, iOS, Android, and Web, with each platform reporting basic (``Bas.'') and advanced (``Adv.'') instruction splits. BiGrounder-4B achieves the best overall average (87.5). The gains are most pronounced on the advanced splits, where instructions carry more implicit intent, which is consistent with our claim that bidirectional self-reinforcement strengthens robustness to instruction variation.

\subsection{UI-Vision}
UI-Vision evaluates grounding under Basic, Functional, and Spatial instruction types. As shown in Table~\ref{tab:ui-vison}, BiGrounder-4B obtains the highest score in every category and the best overall average (46.6), substantially exceeding its Qwen3-VL-4B backbone (26.9). The largest relative improvement appears on Spatial instructions, which require reasoning about relative element positions, highlighting the benefit of learning semantic--spatial alignment from the reverse description task.

\subsection{OSWorld-G-Refine}
Table~\ref{tab:osworld_g_refine} reports the fine-grained results on OSWorld-G-Refine, which categorizes queries into Text Matching, Element Recognition, Layout Understanding, Fine-grained Manipulation, and Refusal. BiGrounder-4B achieves the best Text Matching accuracy and a competitive overall average (69.5), second only to the much larger OpenCUA-32B, while clearly surpassing all open-source models at comparable scale.

\section{Full Per-Benchmark Result Tables}
\label{app:full_results}

Tables~\ref{tab:ui-vison}--\ref{tab:main_results_mmbench} present detailed results across five GUI grounding benchmarks, covering instruction semantics, text and icon targets, professional interfaces, and operating systems. They characterize BiGrounder's performance under diverse semantic and spatial localization requirements.

\par\medskip
\input{appendix/tables/UI_vision}
\par\medskip

% Flush earlier experimental floats before the two table-only benchmark pages.
\clearpage
\input{appendix/sections/full_result_tables}
\clearpage

% Theory follows the two dedicated benchmark-table pages.
\begingroup
\fontsize{9.5}{10.2}\selectfont
\setlength{\abovedisplayskip}{4pt plus 1pt minus 1pt}
\setlength{\belowdisplayskip}{4pt plus 1pt minus 1pt}
\setlength{\abovedisplayshortskip}{0pt plus 1pt}
\setlength{\belowdisplayshortskip}{4pt plus 1pt minus 1pt}
\setlength{\jot}{2pt}
\input{appendix/sections/sampling_analysis}
\par\endgroup

% Prompt templates use one coherent single-column block at the end.
\onecolumn
\raggedbottom
\input{appendix/sections/prompt_templates}

% Restore two columns for the standalone bibliography, if present.
\twocolumn

%% file: appendix/tables/P_N.tex
\begin{table}[t]
\centering
\footnotesize
\setlength{\tabcolsep}{3.2pt}
\begin{tabular*}{\linewidth}{@{\extracolsep{\fill}}lccc}
\toprule
\textbf{Filter} & \textbf{Kept (C / I)} & \textbf{Precision} $\uparrow$ & \textbf{NFR} $\uparrow$ \\
\midrule
\rowcolor{ratioTwenty}
No filter & 273 / 27 & 91.0\% \scriptsize{(273/300)} & 0.0\% \scriptsize{(0/27)} \\
\rowcolor{ratioThirty}
BERTScore $> t$ & 268 / 22 & 92.4\% \scriptsize{(268/290)} & 18.5\% \scriptsize{(5/27)} \\
\rowcolor{ratioThirty}
LLM Judge & 267 / 7 & 97.4\% \scriptsize{(267/274)} & 74.1\% \scriptsize{(20/27)} \\
\rowcolor{ratioFifty}
\textbf{Cycle-Consistency} & \textbf{260 / 2} & \textbf{99.2\%} \scriptsize{(260/262)} & \textbf{92.6\%} \scriptsize{(25/27)} \\
\bottomrule
\end{tabular*}
\caption{Noise filtering on 300 manually annotated reverse instructions, including 273 correct and 27 incorrect ones. \textbf{Precision}: fraction of correct samples in the kept set. \textbf{Noise Filtering Rate (NFR)}: percentage of incorrect samples successfully filtered out.}
\label{tab:P_N}
\end{table}

%% file: appendix/tables/case.tex
\begin{table}[t]
\centering
\scriptsize
\setlength{\tabcolsep}{1.0pt}
\begin{tabular*}{\linewidth}{@{\extracolsep{\fill}}p{0.50\linewidth}ccc}
\toprule
\textbf{Reverse Instruction} & \textbf{BERTScore} & \textbf{Cycle-C.} & \textbf{Reward} \\
\midrule
\rowcolor{ratioTwenty}
Click the star to follow the repository. & 0.86 & \xmark & 0.00 \\
\rowcolor{ratioTwenty}
Click the star to bookmark this webpage. & 0.92 & \checkmark & 0.92 \\
\bottomrule
\end{tabular*}
\caption{Case study showing that Cycle-Consistency filters out high-BERTScore but visually incorrect reverse instructions. GT: ``Click star to save this page.''}
\label{tab:case_reward_hacking}
\end{table}

%% file: appendix/sections/additional_experiments.tex
% Additional experimental tables consolidated from the rebuttal.
\section{Training Analysis and Generalization}
\label{app:additional_experiments}
Tables~\ref{tab:rollout_controls}--\ref{tab:qwen35} report additional rollout-budget controls, instruction-generation ablations, and Qwen3.5 backbone experiments.

\input{appendix/tables/rollout_controls}
\paragraph{Rollout budget and reverse rewards.}
Table~\ref{tab:rollout_controls} separates the gains of BiGrounder from simply sampling more grounding trajectories. Increasing forward-only rollouts from 16 to 20 raises the four-benchmark mean only from 77.3 to 77.5, whereas BiGrounder reaches 79.4 with four reverse and 16 forward trajectories, a gain of 1.9 percentage points at the same total trajectory count. Thus, allocating more rollouts to forward grounding alone does not recover the observed gains. With the same reverse/forward rollout counts, BiGrounder also outperforms Gate-only by 1.1 points (79.4 vs. 78.3), supporting the benefit of graded semantic feedback beyond binary cycle success.
\par\medskip

\input{appendix/tables/generator_ablation}
% \hfill
% \input{appendix/tables/training_seeds}
\paragraph{On-policy generation and bidirectional updates.}
Table~\ref{tab:generator_ablation} compares the three augmentation variants under matched rollout budgets and optimization-step counts. Replacing the frozen Bi-SFT generator with the current policy improves Original accuracy from 68.2 to 68.8 and All accuracy from 59.1 to 60.1, while reducing Gap from 15.9 to 15.1 points. Joint reverse-policy optimization further improves Original to 69.5 and All to 60.5, with Gap reduced to 14.8.

On-policy generation and reverse updates are mutually reinforcing. On-policy sampling refreshes training instructions as the model evolves, while reverse updates help sustain the GUI-understanding capability needed to generate meaningful, target-specific instructions. Because both tasks share model parameters, optimizing the forward objective alone does not explicitly preserve reverse generation, and this capability may weaken as grounding training proceeds. The reverse objective continually reinforces it through grounding-verified semantic feedback, helping maintain useful on-policy instruction exploration. BiGrounder thus couples an evolving source of training instructions with continued optimization of the capability that supports their generation. The observed gains are consistent with this mutually reinforcing design.
\par\medskip

\input{appendix/tables/qwen35}

\noindent\textbf{Dense and MoE backbones.} Table~\ref{tab:qwen35} shows that BiGrounder improves average accuracy on both Qwen3.5 backbones: from 76.3 to 77.7 (+1.4 points) for the dense 9B model, and from 77.7 to 78.9 (+1.2 points) for the 35B-A3B MoE model, supporting the applicability of our training framework across both architectures.

%% file: appendix/tables/rollout_controls.tex
\begin{table*}[!tp]
\centering
\small
\setlength{\tabcolsep}{4pt}
\begin{tabularx}{\textwidth}{l*{7}{>{\centering\arraybackslash}X}}
\toprule
\textbf{Method} & \textbf{Rev.} & \textbf{Fwd.} & \textbf{SS-Pro} & \textbf{SS-V2} & \textbf{OS-G-R} & \textbf{MM-L2} & \textbf{Avg.} \\
\midrule
Forward-only RL ($1\times16$) & 0 & 16 & 63.7 & 92.9 & 67.4 & 85.3 & 77.3 \\
Forward-only RL ($1\times20$) & 0 & 20 & 63.9 & 92.9 & 67.9 & 85.1 & 77.5 \\
BiGrounder Gate-only & 4 & 16 & 65.6 & 92.9 & 68.6 & 86.2 & 78.3 \\
\rowcolor{ratioFifty}
BiGrounder & 4 & 16 & \textbf{66.6} & \textbf{93.9} & \textbf{69.5} & \textbf{87.5} & \textbf{79.4} \\
\bottomrule
\end{tabularx}
\caption{Rollout-budget and reverse-reward controls. Rev./Fwd.: reverse/forward trajectories per target. SS-Pro: ScreenSpot-Pro; SS-V2: ScreenSpot-V2; OS-G-R: OSWorld-G-Refine; MM-L2: MMBench-GUI L2. Scores are accuracy (\%). Avg. is the unweighted mean over these four benchmarks, excluding UI-Vision. The $1\times20$ baseline matches BiGrounder's total trajectory count. Gate-only uses binary reverse rewards.}
\label{tab:rollout_controls}
\end{table*}

%% file: appendix/tables/generator_ablation.tex
\begin{table}[!tbp]
\vspace{0pt}
\centering
\small
\setlength{\tabcolsep}{3pt}
\begin{tabularx}{\linewidth}{l*{4}{>{\centering\arraybackslash}X}}
\toprule
\textbf{Method} & \textbf{Original} & \textbf{Any}$\uparrow$ & \textbf{All}$\uparrow$ & \textbf{Gap}$\downarrow$ \\
\midrule
Forward-only RL & 67.4 & 74.5 & 58.2 & 16.3 \\
Offline aug. & 68.2 & 75.0 & 59.1 & 15.9 \\
Online aug. & 68.8 & 75.2 & 60.1 & 15.1 \\
\rowcolor{ratioFifty}
BiGrounder & \textbf{69.5} & \textbf{75.3} & \textbf{60.5} & \textbf{14.8} \\
\bottomrule
\end{tabularx}
\caption{Instruction-generation and reverse-update ablations on OSWorld-G-Refine-Variants. Original, Any, and All are accuracies (\%); Gap is Any minus All in percentage points. Offline/online augmentation use a frozen Bi-SFT/current-policy instruction generator, respectively, with forward-only updates; BiGrounder also updates the reverse policy. These three methods share rollout budgets and optimization-step counts.}
\label{tab:generator_ablation}
\end{table}

%% file: appendix/tables/qwen35.tex
\begin{table*}[!tp]
\centering
\small
\setlength{\tabcolsep}{5pt}
\begin{tabularx}{\textwidth}{l*{5}{>{\centering\arraybackslash}X}}
\toprule
\textbf{Method} & \textbf{SS-Pro} & \textbf{SS-V2} & \textbf{OS-G-R} & \textbf{MM-L2} & \textbf{Avg.} \\
\midrule
Qwen3.5-9B & 65.2 & 91.8 & 67.7 & 80.4 & 76.3 \\
\rowcolor{ratioTwenty}
\quad + BiGrounder & \textbf{65.7} & \textbf{92.0} & \textbf{68.0} & \textbf{85.1} & \textbf{77.7} \\
\midrule
Qwen3.5-35B-A3B & 64.5 & \textbf{95.2} & 63.6 & 87.5 & 77.7 \\
\rowcolor{ratioTwenty}
\quad + BiGrounder & \textbf{65.3} & 94.7 & \textbf{67.6} & \textbf{88.1} & \textbf{78.9} \\
\bottomrule
\end{tabularx}
\caption{Additional backbone experiments with Bi-SFT on 100k samples followed by bidirectional RL on 10k samples. Scores are accuracy (\%); Avg. is the mean over the four displayed benchmarks. Bold indicates the higher value within each backbone pair.}
\label{tab:qwen35}
\end{table*}

%% file: appendix/sections/method_comparison.tex
\section{Comparison with Related GUI Learning Methods}
\label{app:method_comparison}
Table~\ref{tab:method_comparison} compares BiGrounder with WinClick, UI-Ins, and \emph{Look Where You're Told}. Our contribution is an on-policy bidirectional RL framework for the many-to-one instruction--element mapping problem, connecting instruction exploration with grounding feedback.

\input{appendix/tables/method_comparison}

\paragraph{Relation to prior methods.}
WinClick~\cite{hui2025winclick} learns forward grounding and reverse description prediction through supervised losses. UI-Ins~\cite{chen2025uiinsenhancingguigrounding} uses instruction rewriting as intermediate reasoning with SFT and GRPO; its online reasoning differs from our independent, target-conditioned reverse task. \emph{Look Where You're Told}~\cite{kim2026attentioncycleconsistency} reconstructs instructions from attended visual information and enforces attention consistency, whereas BiGrounder verifies whether generated instructions ground back to the annotated target.

\paragraph{Coupled on-policy reinforcement.}
ScaleCUA~\cite{liu2025scalecua} incorporates GUI-understanding data, including element appearance, functionality, and layout descriptions, into SFT alongside grounding supervision, motivating the use of interface semantics to support localization. BiGrounder extends this principle from SFT to RL: coordinate-conditioned reverse generation continues to train the model to describe GUI elements in language while the forward policy learns grounding. Joint reverse updates help preserve this descriptive capability during grounding optimization, sustaining the source of on-policy instructions rather than leaving it without direct reinforcement. The current policy samples instructions and grounds them back to the target; a positive reverse reward requires both semantic qualification and grounding success. Table~\ref{tab:generator_ablation} demonstrates the grounding benefit: adding reverse updates to online generation raises Original accuracy from 68.8 to 69.5 and All accuracy from 60.1 to 60.5, while reducing Gap from 15.1 to 14.8 points. On-policy exploration and continued GUI understanding thus reinforce one another, using existing annotations and retaining standard forward-only inference.

%% file: appendix/tables/method_comparison.tex
\begin{table*}[!tp]
\centering
\small
\setlength{\tabcolsep}{4pt}
\renewcommand{\arraystretch}{1.25}
\begin{tabularx}{\textwidth}{>{\raggedright\arraybackslash}p{0.27\textwidth}*{5}{>{\centering\arraybackslash}X}}
\toprule
\textbf{Method} & \makecell{\textbf{Generation}\\\textbf{basis}} & \makecell{\textbf{On-policy}\\\textbf{generation}} & \makecell{\textbf{Cycle}\\\textbf{verification}\\\textbf{target}} & \makecell{\textbf{Bidirectional}\\\textbf{RL}} & \textbf{Training} \\
\midrule
WinClick~\cite{hui2025winclick} & Annotations & $\times$ & --- & $\times$ & SFT \\
Look Where You're Told~\cite{kim2026attentioncycleconsistency} & Attention & $\times$ & \makecell{Attention\\distribution} & $\times$ & SFT \\
UI-Ins~\cite{chen2025uiinsenhancingguigrounding} & --- & $\times$ & --- & $\times$ & SFT+GRPO \\
% \rowcolor{ratioFifty}
\textbf{BiGrounder (Ours)} & \textbf{Coordinates} & $\checkmark$ & \textbf{GT location} & $\checkmark$ & \textbf{SFT+GRPO} \\
\bottomrule
\end{tabularx}
\caption{Qualitative comparison of instruction generation and grounding. On-policy generation and bidirectional RL refer to an independent instruction-generation task, excluding UI-Ins's instruction-as-reasoning. Annotations denotes labelled element information without specifying its input encoding. GT: ground truth.}
\label{tab:method_comparison}
\end{table*}

%% file: appendix/tables/UI_vision.tex
\noindent\begin{minipage}{\linewidth}
\centering
\fontsize{8}{9}\selectfont
\renewcommand{\h}[1]{\textbf{\fontsize{7.5}{8.5}\selectfont\textsf{#1}}}
\setlength{\tabcolsep}{1.5pt}
\resizebox{\linewidth}{!}{%
\begin{tabular}{lccccc}
\toprule
\h{Model} & \h{Basic} & \h{Functional} & \h{Spatial}& \h{Avg}\\
\midrule
\rowcolor{gray!15}
\multicolumn{5}{l}{\textit{Proprietary Models}} \\
GPT-4o~\cite{gpt4o}& 1.6 & 1.5 &1.0&\cellcolor{cb}1.4\\
Claude-3.7-Sonnet~\cite{claude37}& 9.5 & 7.7&7.6&\cellcolor{cb}8.3 \\
\midrule
\rowcolor{gray!15}
\multicolumn{5}{l}{\textit{Open-source Models}} \\
Qwen2.5-VL-7B~\cite{bai2025qwen25vltechnicalreport}&1.2& 0.8&0.5&\cellcolor{cb}0.9\\
UGround-V1-7B~\cite{uground}&15.4& 17.1& 6.3 &\cellcolor{cb}12.9\\
OS-Atlas-7B~\cite{osaltas_and_screenspot_v2}&12.2 &11.2 &3.7 &\cellcolor{cb}9.0\\
JEDI-7B~\cite{jedi}& 32.3 &30.5& 12.8 &\cellcolor{cb}24.8 \\
InfiGUI-G1-7B~\cite{liu2025infiguig1} & 36.2 & 31.9 & 11.5 & \cellcolor{cb}26.1 \\
UI-Venus-7B~\cite{ui_venus} &36.1&32.8 &11.9&\cellcolor{cb}26.5\\
Phi-Ground-7B~\cite{phi_ground}&36.8 &37.1 & 7.6 &\cellcolor{cb}27.2\\
Qwen3-VL-4B~\cite{Qwen3-VL}$^*$ & 32.8 &34.2 &14.7 &\cellcolor{cb}26.9\\
MAI-UI-8B~\cite{zhou2025maiuitechnicalreportrealworld} &\underline{51.7} &\underline{49.6} &22.5 &\cellcolor{cb}\underline{40.7} \\
\midrule
UI-TARS-72B~\cite{ui-tars}& 31.4 &30.5 & 14.7 & \cellcolor{cb}25.5\\
UI-Venus-72B~\cite{ui_venus}& 45.6 & 42.3 & \underline{23.7}&\cellcolor{cb}36.8\\
\midrule
\rowcolor{gray!15}
\multicolumn{5}{l}{\textit{Ours}} \\
BiGrounder-4B& \textbf{55.9} & \textbf{52.8} & \textbf{32.3}&\cellcolor{cb}\textbf{46.6}\\
\bottomrule
\end{tabular}
}
\captionof{table}{Performance comparison on the \textbf{UI-Vision} benchmark. The best results are highlighted in \textbf{bold}, and the second-best results are \underline{underlined}.}
\label{tab:ui-vison}
\end{minipage}

%% file: appendix/sections/full_result_tables.tex
% Two table-only pages within the paper's two-column layout.
% Measure each complete page once and scale down only if it exceeds the page height.
\newsavebox{\BGHFirstPage}
\newsavebox{\BGHSecondPage}
\newcommand{\BGHFitPage}[1]{%
  \ifdim\dimexpr\ht#1+\dp#1\relax>\textheight
    \resizebox*{!}{\textheight}{\usebox{#1}}%
  \else
    \usebox{#1}%
  \fi
}

\begin{table*}[!p]
\centering
\begin{lrbox}{\BGHFirstPage}
\begin{minipage}{\textwidth}
\par\centering
\begin{minipage}{0.95\linewidth}
\input{appendix/tables/os-world-g-refined}
\end{minipage}
\par\vspace{12pt}
\begin{minipage}{0.80\linewidth}
\input{appendix/tables/ssv2}
\end{minipage}
\end{minipage}
\end{lrbox}
\typeout{BG-H-FIRST-PAGE-HEIGHT=\the\dimexpr\ht\BGHFirstPage+\dp\BGHFirstPage\relax}
\BGHFitPage{\BGHFirstPage}
\end{table*}

\begin{table*}[!p]
\centering
\begin{lrbox}{\BGHSecondPage}
\begin{minipage}{\textwidth}
\centering
\begin{minipage}{0.95\linewidth}
\input{appendix/tables/sspro}
\end{minipage}
\par\vspace{14pt}
\begin{minipage}{0.95\linewidth}
\input{appendix/tables/mmbench}
\end{minipage}
\end{minipage}
\end{lrbox}
\typeout{BG-H-SECOND-PAGE-HEIGHT=\the\dimexpr\ht\BGHSecondPage+\dp\BGHSecondPage\relax}
\BGHFitPage{\BGHSecondPage}
\end{table*}

%% file: appendix/tables/os-world-g-refined.tex
\noindent\begin{minipage}{\linewidth}
    \centering
    \setlength{\tabcolsep}{2.5pt}
    \footnotesize
    \resizebox{\linewidth}{!}{%
    \begin{tabular}{lcccccc}
        \toprule
        \h{Model} & \makecell[c]{\h{Text}\\\h{Matching}} & \makecell[c]{\h{Element}\\\h{Recognition}} & \makecell[c]{\h{Layout}\\\h{Understanding}} & \makecell[c]{\h{Fine-grained}\\\h{Manipulation}}& \h{Refusal} & \cellcolor{white}\h{Avg} \\
        \midrule
        \rowcolor{gray!15}
        \multicolumn{7}{l}{\textit{Open-Source Models}} \\
        Jedi-3B~\cite{jedi} & - & - & - & - & - &\cellcolor{cb}61.0 \\
        Jedi-7B~\cite{jedi} & - & - & - & - & - & \cellcolor{cb}63.8 \\
        UI-TARS-1.5-7B~\cite{ui-tars-15} & 52.6 & 75.4 & 72.4 & \underline{66.7} & 0.0 & \cellcolor{cb}64.2\\
        Qwen3-VL-4B~\cite{Qwen3-VL}$^*$ & \underline{78.2} &71.6 &76.5 &55.2 & 1.9 &\cellcolor{cb}65.3\\
        GTA1-7B~\cite{GTA1} & 63.2 & \textbf{82.1} & 74.2  & \textbf{70.5} & 0.0 & \cellcolor{cb}67.7 \\
        GUI-Owl-1.5-4B-Instruct~\cite{gui-owl-1.5} & 73.1 & 71.8 & 72.9 & \textbf{74.1} & 29.6 & \cellcolor{cb}68.4 \\
        \midrule
        Qwen3-VL-32B~\cite{Qwen3-VL} & 77.4 & 73.6 & 76.3 & 57.7 & - & \cellcolor{cb}69.0 \\
        OpenCUA-32B~\cite{opencua} & 63.2 & 79.9 & \textbf{84.9} & 62.1 & 7.4 & \cellcolor{cb}\textbf{70.2} \\
        \midrule
        \rowcolor{gray!15}
        \multicolumn{7}{l}{\textit{Ours}} \\
        BiGrounder-4B & \textbf{77.4}  & \underline{76.7} & \underline{77.1} & 56.4 & 0.0 & \cellcolor{cb}\underline{69.5} \\
        \bottomrule
    \end{tabular}
    }
    \captionof{table}{Performance comparison on the \textbf{OSWorld-G-Refine} benchmark. The best results are highlighted in \textbf{bold}, and the second-best results are \underline{underlined}.}
    \label{tab:osworld_g_refine}
\end{minipage}

%% file: appendix/tables/ssv2.tex
\noindent\begin{minipage}{\linewidth}
    \centering
    \footnotesize
    \setlength{\tabcolsep}{5.0pt}
    \resizebox{\linewidth}{!}{%
        \begin{tabular}{lccccccc}
        \toprule
        % --- Top group ---
        \multirow{2}{*}{\h{Model}}
        % --- Second-level group ---
        & \multicolumn{2}{c}{\h{Mobile}} &
          \multicolumn{2}{c}{\h{Desktop}} &
          \multicolumn{2}{c}{\h{Web}} &
          \multirow{2}{*}{\h{Avg.}}  \\
        \cmidrule(lr){2-3} \cmidrule(lr){4-5} \cmidrule(lr){6-7}

        % --- Column titles ---
        & \h{Text} & \h{Icon.} & \h{Text} & \h{Icon.} & \h{Text} & \h{Icon.} &  \\
        \midrule
        \rowcolor{gray!15}
        \multicolumn{8}{l}{\textit{Open-Source Models}} \\
        % --- 7B Models (Baselines) ---
        Qwen3-VL-2B~\cite{Qwen3-VL} &95.5 &82.0 &95.4 &73.6 &89.7 &76.4 &\cellcolor{cb}86.7 \\
        Phi-ground~\cite{phi_ground}
            & 90.2 & 76.4 & 93.6 & 75.9 & \underline{96.5} & 62.0
            & \cellcolor{cb}83.8
            \\
        OS-Atlas-7B~\cite{osaltas_and_screenspot_v2}
            & 95.2 & 75.8 & 90.7 & 63.6 & 90.6 & 77.3
            & \cellcolor{cb}85.1
           \\
        UGround-v1-7B~\cite{uground}
            & 83.6 & 90.5 & 85.8 & 86.3 & 95.5 & 83.2
            & \cellcolor{cb}87.7
            \\
        UI-Tars-1.5-7B~\cite{ui-tars-15}
            & 92.2 & 81.5 & 91.0 & 84.2 & 95.5 & 84.5
            & \cellcolor{cb}90.3
            \\
        SE-GUI-7B~\cite{SE_GUI}
            & - & - &
              - & - &
              - & - &
              \cellcolor{cb}68.9
              \\
        UI-TARS-7B~\cite{ui-tars}
            & 96.9 & 89.1 & 95.4 & 85.0 & 93.6 & 85.2
            & \cellcolor{cb}91.6
             \\
        Qwen3-VL-8B~\cite{Qwen3-VL} &97.9 &84.8 &95.9 &87.9 &95.7 &83.7 &\cellcolor{cb}91.7\\
        GUI-Actor-7B~\cite{wu2025gui}
            & 97.6 & 88.2 & 96.9 & 85.7 & 93.2 & 86.7
            & \cellcolor{cb}92.1
             \\
        OpenCUA-7B~\cite{opencua}
            & - & - & - & - & - & -
            & \cellcolor{cb}92.3
            \\
        GTA1-7B~\cite{GTA1}
            & \underline{99.0} & 88.6 & 94.9 & \underline{89.3} & 92.3 & 86.7
            & \cellcolor{cb}92.4
             \\
        GUI-Owl-7B~\cite{ye2025mobileagentv3fundamentalagentsgui}
            & \underline{99.0} & \textbf{92.4} & 96.9 & 85.0 & 93.6 & 85.2 &
            \cellcolor{cb}92.8 \\
        GUI-G$^2$-7B~\cite{GUI_G2}
            & 98.3 & \underline{91.9} &
              95.4 & \underline{89.3} &
              94.0 & 87.7 &
              \cellcolor{cb}93.3
              \\
        InfiGUI-G1-7B~\cite{liu2025infiguig1}
            & \underline{99.0} & \underline{91.9} &
              94.3 & 82.1 &
              \textbf{97.9} & 89.2 &
              \cellcolor{cb}\underline{93.5}
              \\
        MAI-UI-2B~\cite{zhou2025maiuitechnicalreportrealworld}& \textbf{99.3} &87.2 &97.4 &88.6 &94.0 &84.7 &\cellcolor{cb}92.5\\
        \midrule
        % --- Larger Models (Baselines) ---
        Qwen3-VL-32B~\cite{Qwen3-VL} & 96.2 &90.0 &97.4 &85.0 &95.7 & \underline{89.7} &\cellcolor{cb}93.0\\
        GUI-Owl-32B~\cite{ye2025mobileagentv3fundamentalagentsgui}
            & 98.6 & 90.0 & \underline{97.9} & 87.8 & 94.4 & 86.7 &
            \cellcolor{cb}93.2 \\
        OpenCUA-32B~\cite{opencua}
            & - & - & - & - & - & -
            & \cellcolor{cb}93.4
            \\
        \midrule
        \rowcolor{gray!15}
        \multicolumn{8}{l}{\textit{Ours}} \\
        % --- Our Models ---
        BiGrounder-4B & 97.8 & 90.1 & \textbf{98.1} & \textbf{92.5} & 94.2 & \textbf{90.7} & \cellcolor{cb}\textbf{93.9} \\
        \bottomrule
        \end{tabular}
    }
    \captionof{table}{Performance comparison on \textbf{ScreenSpot-V2}. The best results are highlighted in \textbf{bold}, and the second-best results are \underline{underlined}.}
    \label{tab:ssv2}
\end{minipage}

%% file: appendix/tables/sspro.tex
\noindent\begin{minipage}{\linewidth}
    \centering
    \footnotesize
    \setlength{\tabcolsep}{2.0pt}
    \resizebox{\linewidth}{!}{%
        \begin{tabular}{lccccccccccccccc}
        \toprule
        \multirow{2}{*}{\h{Model}} &
        \multicolumn{2}{c}{\h{CAD}} &
        \multicolumn{2}{c}{\h{Dev.}} &
        \multicolumn{2}{c}{\h{Creative}} &
        \multicolumn{2}{c}{\h{Scientific}} &
        \multicolumn{2}{c}{\h{Office}} &
        \multicolumn{2}{c}{\h{OS}} &
        \multirow{2}{*}{\h{Avg.}} \\
        \cmidrule(lr){2-3}
        \cmidrule(lr){4-5}
        \cmidrule(lr){6-7}
        \cmidrule(lr){8-9}
        \cmidrule(lr){10-11}
        \cmidrule(lr){12-13}
        & \h{Text} & \h{Icon} & \h{Text} & \h{Icon} & \h{Text} & \h{Icon} & \h{Text} & \h{Icon} & \h{Text} & \h{Icon} & \h{Text} & \h{Icon} & \\
        \midrule
        \rowcolor{gray!15}
        \multicolumn{14}{l}{\textit{Proprietary Models}} \\
        GPT-4o~\cite{gpt4o}
            & 2.0  & 0.0  & 1.3  & 0.0  & 1.0  & 0.0  & 2.1  & 0.0
            & 1.1  & 0.0  & 0.0  & 0.0  & \cellcolor{cb}0.8 \\

        Gemini-3-Pro~\cite{gemini3pro}
            & -    & -    & -    & -    & -    & -    & -    & -
            & -    & -    & -    & -    & \cellcolor{cb}\textbf{72.7} \\
        \midrule
        \rowcolor{gray!15}
        \multicolumn{14}{l}{\textit{Open-Source Models}} \\
        UI-R1-3B~\cite{ui_r1}
            & 11.2 & 6.3  & 22.7 & 4.1  & 27.3 & 3.5  & 42.4 & 11.8 & 32.2 & 11.3 & 13.1 & 4.5  & \cellcolor{cb}17.8 \\
        GUI-R1-7B~\cite{luo2025guir1generalistr1style}
        & 23.9 & 6.3 & 49.4 & 4.8 & 38.9 & 8.4 & 55.6 & 11.8 & 58.7 & 26.4 & 42.1 & 16.9 & \cellcolor{cb}31.0 \\
        Phi-Ground-7B~\cite{phi_ground}
            & 26.9 & 17.2 & 70.8 & 16.7 & 56.6 & 13.3 & 58.0 & 29.1 & 76.4 & 44.0 & 55.1 & 25.8 & \cellcolor{cb}43.2\\
        GUI-Actor-7B~\cite{wu2025gui}
            & 47.7 & 9.4  & 59.1 & 15.9 & 59.6 & 16.1 & 70.1 & 25.5
            & 69.5 & 41.5 & 55.1 & 19.1 & \cellcolor{cb}44.6 \\
        SE-GUI-7B~\cite{SE_GUI}
            & 51.3 & 14.1 & 68.2 & 19.3 & 57.6 & 9.1  & 75.0 & 28.2
            & 78.5 & 43.4 & 49.5 & 25.8
            & \cellcolor{cb}47.3 \\
        GUI-G$^2$-7B~\cite{GUI_G2}
            & 55.8 & 12.5 & 68.8 & 17.2 & 57.1 & 15.4 & 77.1 & 24.5
            & 74.0 & 32.7 & 57.9 & 21.3 & \cellcolor{cb}47.5 \\
        ScaleCUA-7B~\cite{liu2025scalecua}
            & 54.8 & 21.9 & 66.2 & 20.7 & 56.6 & 20.3 & 77.1 & 24.5
        & 74.0 & 45.3 & 49.5 & 18.0 & \cellcolor{cb}47.9 \\
        UI-TARS-1.5-7B~\cite{ui-tars-15}
            & -    & -    & -    & -    & -    & -    & -    & -
            & -    & -    & -    & -    & \cellcolor{cb}49.6 \\
        OpenCUA-7B~\cite{opencua}
            & -    & -    & -    & -    & -    & -    & -    & -
            & -    & -    & -    & -    & \cellcolor{cb}50.0 \\
        InfiGUI-G1-7B~\cite{liu2025infiguig1}
            & 57.4 & 23.4 & 74.7 & 24.1 & 64.6 & 18.2 & 80.6 & 31.8
            & 75.7 & 39.6 & 57.0 & 29.2 & \cellcolor{cb}51.9 \\
        UI-Ins-7B~\cite{chen2025uiinsenhancingguigrounding} & 60.9 & 20.3 & 75.3 & 18.6 & 65.2 & 18.9 & \underline{81.3} & 29.1 & 79.7 & 37.7 & 57.0 & 25.8 & \cellcolor{cb}52.2 \\
        GUI-Owl-7B~\cite{ye2025mobileagentv3fundamentalagentsgui}
            & 64.5 & 21.9 & 76.6 & 31.0 & 59.6 & 27.3 & 79.1 & \underline{37.3} & 77.4 & 39.6 & 59.8 & 33.7 & \cellcolor{cb}54.9 \\
        % \midrule
        Qwen3-VL-4B~\cite{Qwen3-VL}$^*$ &59.4 &20.3 &\underline{81.8} &33.8 &\underline{71.2} &23.8 &80.6 &34.5 &\textbf{88.8} &47.2 &\textbf{78.5} &41.6 &\cellcolor{cb}59.3 \\
        StepGUI-4B~\cite{step-gui}
        & -    & -    & -    & -    & -    & -    & -    & -
        & -    & -    & -    & -  &\cellcolor{cb}60.0 \\
        MAI-UI-8B~\cite{zhou2025maiuitechnicalreportrealworld} & \textbf{72.6} & \underline{35.9} & \textbf{83.8} & \underline{52.4} & \textbf{76.3} & \underline{33.6} & 79.9 & \underline{37.3} & \underline{88.7} & \textbf{60.4} & \underline{76.6} & \underline{49.4} & \cellcolor{cb}65.8 \\
        \midrule

        \rowcolor{gray!15}
        \multicolumn{14}{l}{\textit{Ours}} \\
        BiGrounder-4B &\underline{68.5} &\textbf{45.3} &81.2 &\textbf{62.0} &\underline{71.2} &\textbf{37.0} &\textbf{84.0} &\textbf{47.3} &86.4 &\underline{52.8} &74.8 &\textbf{54.7} &\cellcolor{cb}\underline{66.6} \\
        \bottomrule
        \end{tabular}
    }
    \captionof{table}{Performance comparison on the \textbf{ScreenSpot-Pro} benchmark. The best results are highlighted in \textbf{bold}, and the second-best results are \underline{underlined}.}
    \label{tab:screenspot_pro}
\end{minipage}

%% file: appendix/tables/mmbench.tex
\noindent\begin{minipage}{\linewidth}
    \centering
    \small
    \setlength{\tabcolsep}{1.6pt}
    \resizebox{\linewidth}{!}{%
    \begin{tabular}{l *{12}{c} c}
        \toprule
        \multirow{2}{*}{\h{Model}} &
        \multicolumn{2}{c}{\h{Windows}} &
        \multicolumn{2}{c}{\h{MacOS}} &
        \multicolumn{2}{c}{\h{Linux}} &
        \multicolumn{2}{c}{\h{iOS}} &
        \multicolumn{2}{c}{\h{Android}} &
        \multicolumn{2}{c}{\h{Web}} &
        \multirow{2}{*}{\h{Avg.}} \\
        \cmidrule(lr){2-3}
        \cmidrule(lr){4-5}
        \cmidrule(lr){6-7}
        \cmidrule(lr){8-9}
        \cmidrule(lr){10-11}
        \cmidrule(lr){12-13}
        & Bas. & Adv. & Bas. & Adv. & Bas. & Adv. & Bas. & Adv. & Bas. & Adv. & Bas. & Adv. & \\
        \midrule
        \rowcolor{gray!15}
        \multicolumn{14}{l}{\textit{Proprietary Models}} \\
        % General Models
        GPT-4o~\cite{gpt4o}
            & 1.5 & 1.1 & 8.7 & 4.3 & 1.1 & 1.0 & 5.1 & 3.3 & 2.5 & 1.4 & 3.2 & 2.9
            & \cellcolor{cb}2.9 \\
        Claude-3.7-Sonnet~\cite{claude37}
            & 1.5 & 0.7 & 12.5 & 7.5 & 1.1 & 0.0 & 13.7 & 10.6 & 1.4 & 1.4 & 3.2 & 2.3
            & \cellcolor{cb}4.7 \\
        \midrule
        \rowcolor{gray!15}
        \multicolumn{14}{l}{\textit{Open-Source Models}} \\
        OS-Atlas-7B~\cite{osaltas_and_screenspot_v2}
            & 36.9 & 18.8 & 44.4 & 21.7 & 31.4 & 13.3 & 74.8 & 48.8 & 69.6 & 46.8 & 61.3 & 35.4
            & \cellcolor{cb}41.4 \\
        Aguvis-7B~\cite{aguvis}
            & 37.3 & 21.7 & 48.1 & 33.3 & 33.5 & 25.0 & 67.5 & 65.2 & 61.0 & 51.0 & 61.6 & 45.5
            & \cellcolor{cb}45.7 \\
        UI-TARS-1.5-7B~\cite{ui-tars-15}
            & 68.3 & 39.0 & 69.0 & 44.5 & 64.4 & 37.8 & 88.5 & 69.4 & 90.5 & 69.3 & 81.0 & 56.5
            & \cellcolor{cb}64.3 \\
        Qwen2.5-VL-7B~\cite{bai2025qwen25vltechnicalreport}$^*$
            & 69.1 & 42.7 & 70.5 & 51.4 & 74.4 & 30.8 & 86.9 & 85.5 & 85.7 & 85.1 & 78.9 & 74.6
            & \cellcolor{cb}69.9 \\
        SE-GUI-7B~\cite{SE_GUI}
            & - & - & - & - & - & - & - & - & - & - & - & -
            & \cellcolor{cb}34.5 \\
        ScaleCUA-7B~\cite{liu2025scalecua}
            & 78.6 & 54.0 & 82.3 & 58.7 & 74.4 & 56.6 & 94.3 & 81.8 & \underline{96.1} & 81.1 & 92.6 & 73.1 & \cellcolor{cb}78.2 \\
        Qwen3-VL-4B~\cite{Qwen3-VL}$^*$
            & \underline{88.6} & 61.8 & 85.5 & 69.1 & 74.9 & 53.1 & \underline{95.2} & 82.4 & 95.5 & 84.4 & \textbf{96.8} & 72.1 &\cellcolor{cb}81.2\\
        GUI-Owl-7B~\cite{ye2025mobileagentv3fundamentalagentsgui}
            & 86.4 & 61.8 & 81.7 & 64.5 & 74.4 & 61.7
            & 94.9 & 83.0 & 95.8 & 83.7 & 93.2 & 72.7
            & \cellcolor{cb}80.5 \\
        InfiGUI-G1-7B~\cite{liu2025infiguig1}
            & 82.7 & 61.8 & 83.8 & 63.9 & 72.3 & 52.0 & 94.9 & 89.4 & 95.2 & 85.6 & 93.5 & 76.3
            & \cellcolor{cb}80.8 \\
        UI-Ins-7B~\cite{chen2025uiinsenhancingguigrounding}
            & 82.7 & 64.7 & 87.2 & \underline{75.1} & 71.7 & 51.5 & 94.9 & \underline{89.7} & 95.8 & \underline{89.0} & 93.2 & \underline{80.8} & \cellcolor{cb}83.1 \\

        % > 7B
        \midrule
        InternVL3-78B~\cite{internvl3}
            & 70.1 & 42.6 & 75.7 & 52.3 & 59.2 & 41.3 & 93.6 & 80.6 & 92.7 & 78.6 & 90.7 & 65.9
            & \cellcolor{cb}72.2 \\
        UI-TARS-DPO-72B~\cite{ui-tars}
            & 78.6 & 51.8 & 80.3 & 62.7 & 68.6 & 51.5 & 90.8 & 81.2 & 93.0 & 80.0 & 88.1 & 68.5
            & \cellcolor{cb}74.3 \\
        GUI-Owl-32B~\cite{ye2025mobileagentv3fundamentalagentsgui}
            & 85.6 & 65.1 & 84.9 & 67.1 & 77.0 & \underline{63.3}
            & \underline{95.2} & 85.5 & \underline{96.1} & 87.0 & \underline{95.5} & \underline{80.8}
            & \cellcolor{cb}83.0 \\
        Qwen3-VL-32B~\cite{Qwen3-VL}
            & \textbf{93.4} & \underline{71.3} & \textbf{92.8} & 74.3 & \underline{78.0} & 56.1 & \textbf{95.5} & 88.8 & \textbf{97.2} & 88.5 & 92.6 & 78.6 &\cellcolor{cb}\underline{85.3}\\

        \midrule
        \rowcolor{gray!15}
        \multicolumn{14}{l}{\textit{Ours}} \\
        BiGrounder-4B
            & 85.5 & \textbf{78.1} & \underline{87.3} & \textbf{81.0} & \textbf{88.6} & \textbf{76.0} & 93.1 & \textbf{92.0} & 93.5 & \textbf{93.5} & 92.8 & \textbf{86.8}
            & \cellcolor{cb}\textbf{87.5} \\
        \bottomrule
    \end{tabular}
    }
    \captionof{table}{Performance comparison on the \textbf{MMBench-GUI L2} benchmark. ``Bas.'' and ``Adv.'' denote the basic and advanced instruction splits, respectively. The best results are highlighted in \textbf{bold}, and the second-best results are \underline{underlined}.}
    \label{tab:main_results_mmbench}
\end{minipage}

%% file: appendix/sections/sampling_analysis.tex
\section{A Policy-Gradient Analysis of Multi-Instruction and On-Policy Sampling}
\label{app:sampling_analysis}
We analyze two aspects of BiGrounder's reinforcement-learning design: allocating grounding rollouts across multiple instructions, and refreshing instructions with the evolving reverse policy. The first reduces sampling variance for a fixed expected-reward objective. The second avoids instruction-distribution mismatch when estimating the current objective, yielding a conditional mean-squared-error advantage over a fixed instruction source.

\subsection{Grounding as Expected-Reward Optimization}
Fix a screenshot $s$, target $o^*$, and training iteration $t$. Let $q_t$ denote the instruction distribution induced by the current reverse policy, including the semantic check and ground-truth fallback. We idealize these instructions as target-preserving; passing a semantic threshold alone does not guarantee this assumption. The forward grounding policy $\pi_\theta(\tau\mid s,x)$ generates a token trajectory $\tau$, evaluated by the parameter-independent grounding reward $R(\tau,o^*)$. Its local objective is
\begin{equation}
J_t^{\mathrm{fwd}}(\theta)
=\mathbb E_{x\sim q_t}\mathbb E_{\tau\sim\pi_\theta(\cdot\mid s,x)}
\left[R(\tau,o^*)\right].
\label{eq:sampling_rl_objective}
\end{equation}
The distribution $q_t$ is held fixed in this forward-policy derivative, even though the generator and grounder share parameters in implementation. Thus, we analyze the forward contribution to the update, rather than the full derivative through instruction generation.

Write $V_\theta(x)=\mathbb E_{\tau\sim\pi_\theta}[R(\tau,o^*)]$ and let $b_t(x)$ be a baseline independent of the sampled trajectory. The score-function policy-gradient sample at $\theta_t$ is
\begin{equation}
Z_t(x,\tau)=\bigl(R(\tau,o^*)-b_t(x)\bigr)
\left.\nabla_\theta\log\pi_\theta(\tau\mid s,x)\right|_{\theta_t}.
\label{eq:sampling_reinforce}
\end{equation}
For an autoregressive model, the trajectory score is the sum of token log-probability gradients. Under the usual differentiation-under-expectation conditions,
\begin{align}
\mathbb E[Z_t\mid x]
&=\left.\nabla_\theta V_\theta(x)\right|_{\theta_t}
 =:g_t(x),\label{eq:sampling_conditional_gradient}\\
\mathbb E_{x\sim q_t,\tau\sim\pi_{\theta_t}}[Z_t]
&=\left.\nabla_\theta J_t^{\mathrm{fwd}}(\theta)\right|_{\theta_t}
 =:\mu_t.\label{eq:sampling_mean_gradient}
\end{align}
Indeed, differentiating the trajectory probabilities gives the reward term, while the baseline contributes $b_t(x)\nabla_\theta\sum_\tau\pi_\theta(\tau\mid s,x)=0$. This establishes conditional unbiasedness directly from the reinforcement-learning objective.

\subsection{Why Multiple Instructions Reduce Gradient Variance}
Assume finite second moments and define total variance for a vector $U$ as $\operatorname{Var}(U)=\mathbb E\|U-\mathbb E U\|_2^2$. Let
\begin{align}
B_t&=\operatorname{Var}_{x\sim q_t}\!\left(g_t(x)\right),\label{eq:sampling_between}\\
W_t&=\mathbb E_{x\sim q_t}\!\left[\operatorname{Var}(Z_t\mid x)\right].\label{eq:sampling_within}
\end{align}
Here $B_t$ captures differences in expected policy gradients across valid expressions of the same target; $W_t$ captures trajectory noise for a given expression. Semantically equivalent instructions need not produce identical gradients in a language-conditioned policy.

\begin{samplingproposition}
\label{prop:instruction_sampling}
Sample $M$ instructions independently from $q_t$ and $K$ conditionally independent grounding trajectories per instruction, with independent instruction groups. Then
\begin{equation}
\hat g_{M,K}=\frac{1}{M}\sum_{i=1}^{M}\bar Z_i,
\qquad
\bar Z_i=\frac{1}{K}\sum_{k=1}^{K}Z_t(x_i,\tau_{i,k})
\end{equation}
is unbiased for $\mu_t$. At a forward-rollout budget $N=MK$, its mean-squared error satisfies
\begin{equation}
\mathbb E\|\hat g_{M,K}-\mu_t\|_2^2
=\operatorname{Var}(\hat g_{M,K})
=\frac{B_t}{M}+\frac{W_t}{N}.
\label{eq:instruction_sampling_variance}
\end{equation}
\end{samplingproposition}

\noindent\textit{Proof.}
Equation~\ref{eq:sampling_conditional_gradient} gives $\mathbb E[\bar Z_i\mid x_i]=g_t(x_i)$, and conditional independence gives $\operatorname{Var}(\bar Z_i\mid x_i)=\operatorname{Var}(Z_t\mid x_i)/K$. Therefore,
\begin{equation}
\operatorname{Var}(\bar Z_i)=B_t+\frac{W_t}{K}
\end{equation}
by total variance. Averaging $M$ independent groups divides this variance by $M$. The tower property gives $\mathbb E[\hat g_{M,K}]=\mu_t$; hence variance equals mean-squared error.\hfill$\square$

For one instruction with 16 rollouts versus four independently sampled instructions with four rollouts each,
\begin{align}
\operatorname{Var}(\hat g_{1,16})&=B_t+\frac{W_t}{16},\\
\operatorname{Var}(\hat g_{4,4})&=\frac{B_t}{4}+\frac{W_t}{16}.
\end{align}
Repeated rollouts of one instruction better estimate that instruction's gradient, but do not average away the choice of wording. Four instructions reduce the between-expression component by 75\%, decreasing total mean-squared error by $3B_t/4$ when $B_t>0$. This comparison holds $q_t$ fixed and counts only the same 16 forward grounding rollouts; it does not equate total generation cost.

\paragraph{Implication for expected-reward improvement.}
If $J_t^{\mathrm{fwd}}$ has an $L$-Lipschitz gradient along the update and $\theta^+=\theta_t+\eta\hat g_{M,K}$, smoothness and unbiasedness give
\begin{align}
\mathbb E[J_t^{\mathrm{fwd}}(\theta^+)]
&\ge J_t^{\mathrm{fwd}}(\theta_t)+\eta\|\mu_t\|_2^2\notag\\
&\quad-\frac{L\eta^2}{2}\left(\|\mu_t\|_2^2+\frac{B_t}{M}+\frac{W_t}{N}\right).
\label{eq:sampling_reward_improvement}
\end{align}
To obtain this inequality, apply the smoothness lower bound to the update and use $\mathbb E\|\hat g_{M,K}\|_2^2=\|\mu_t\|_2^2+\operatorname{Var}(\hat g_{M,K})$. At the same step size and 16 forward rollouts, using four instructions raises this lower bound by $3L\eta^2B_t/8$. This connects variance reduction to a larger lower bound on expected RL reward improvement, rather than guaranteeing the ordering of realized rewards.

\subsection{When On-Policy Instructions Improve Gradient Estimation}
Let $q_0$ be a frozen instruction source, such as a frozen SFT generator or the sampling distribution over a fixed library. All grounding trajectories below are still freshly generated by $\pi_{\theta_t}$. Thus, we isolate instruction-source mismatch rather than stale grounding trajectories. Both estimators use $M$ independent instruction draws and $K$ grounding rollouts per draw; for a fixed library, draws are with replacement and the analysis conditions on that library.

The unweighted fixed-source estimator has expectation $\mu_{0,t}=\mathbb E_{x\sim q_0}[g_t(x)]$. Relative to the current objective in Eq.~\ref{eq:sampling_rl_objective}, its bias is
\begin{equation}
\delta_t=\mu_{0,t}-\mu_t
=\mathbb E_{q_0}[g_t(x)]-\mathbb E_{q_t}[g_t(x)].
\label{eq:sampling_distribution_bias}
\end{equation}
Define $B_{0,t}=\operatorname{Var}_{q_0}(g_t(x))$ and $W_{0,t}=\mathbb E_{q_0}[\operatorname{Var}(Z_t\mid x)]$, both evaluated at the same current grounding policy.

\begin{samplingproposition}
\label{prop:onpolicy_instruction_mse}
Under the preceding assumptions, the on-policy-source and fixed-source estimators have errors
\begin{align}
\mathcal E_{\mathrm{on}}
&:=\mathbb E\|\hat g_{\mathrm{on}}-\mu_t\|_2^2
=\frac{B_t}{M}+\frac{W_t}{N},\\
\mathcal E_{\mathrm{fixed}}
&:=\mathbb E\|\hat g_{\mathrm{fixed}}-\mu_t\|_2^2\notag\\
&=\frac{B_{0,t}}{M}+\frac{W_{0,t}}{N}+\|\delta_t\|_2^2.
\label{eq:sampling_onpolicy_mse}
\end{align}
Consequently, on-policy instruction sampling has strictly smaller mean-squared error whenever
\begin{equation}
\|\delta_t\|_2^2>
\frac{B_t-B_{0,t}}{M}+\frac{W_t-W_{0,t}}{N}.
\label{eq:sampling_onpolicy_condition}
\end{equation}
\end{samplingproposition}

\noindent\textit{Proof.}
Apply Proposition~\ref{prop:instruction_sampling} under each instruction distribution. For the fixed source, write $\hat g_{\mathrm{fixed}}-\mu_t=(\hat g_{\mathrm{fixed}}-\mu_{0,t})+\delta_t$. The first term has zero mean, so its cross term with $\delta_t$ vanishes. The bias--variance decomposition gives Eq.~\ref{eq:sampling_onpolicy_mse}; subtraction gives Eq.~\ref{eq:sampling_onpolicy_condition}.\hfill$\square$

For a fixed iteration and fixed $K$, increasing $M$ drives $\mathcal E_{\mathrm{on}}$ to zero, whereas $\mathcal E_{\mathrm{fixed}}$ approaches $\|\delta_t\|_2^2$. If $\delta_t\ne0$, on-policy sampling therefore has smaller error for sufficiently large matched budgets. More grounding rollouts cannot remove this distribution-mismatch term. However, different instruction distributions need not yield different mean gradients, and a well-matched fixed library can have zero bias or lower variance. This is a conditional advantage for tracking the current generator-defined objective, not proof that this objective is optimal for held-out user instructions.

\paragraph{Importance correction and coverage.}
If $q_t$ is absolutely continuous with respect to $q_0$, instruction-level importance weights $w_t(x)=q_t(x)/q_0(x)$ recover the current gradient:
\begin{equation}
\mathbb E_{q_0,\pi_{\theta_t}}[w_t(x)Z_t(x,\tau)]=\mu_t.
\end{equation}
The bias is therefore avoidable in principle, but correction requires coverage and accurate instruction probabilities. Let $\hat g_{\mathrm{IS}}=M^{-1}\sum_i w_t(x_i)\bar Z_i$ and $\chi^2(q_t\|q_0)=\mathbb E_{q_0}[(w_t-1)^2]$. If $\mathbb E[\|\bar Z_i\|_2^2\mid x]\le C^2$ uniformly, the corrected estimator obeys
\begin{align}
\operatorname{Var}(\hat g_{\mathrm{IS}})
&\le\frac{C^2}{M}\mathbb E_{q_0}[w_t(x)^2]\notag\\
&=\frac{C^2}{M}\bigl(1+\chi^2(q_t\|q_0)\bigr).
\label{eq:sampling_importance_bound}
\end{align}
This follows by bounding variance by the second moment of an independent weighted group and using $\mathbb E_{q_0}[w_t]=1$. On-policy sampling sets the weights to one and removes this mismatch factor from the bound. The bound does not imply that every importance-weighted estimator has higher actual variance. If a finite library omits instructions with positive current probability, exact importance correction cannot recover their contributions.

\subsection{Connection to GRPO and Bidirectional Updates}
In language-model GRPO, responses are sampled from a recent policy snapshot, rewarded, and used in a clipped, importance-ratio update before the sampling policy is refreshed~\cite{grpo}. Fixed question prompts are compatible with on-policy response sampling. Accordingly, a fixed instruction library does \emph{not} by itself make the forward grounding trajectories off-policy.

BiGrounder adds a second policy-controlled trajectory: the instruction itself is the output of the reverse policy. To distinguish the two roles of shared parameters, write the raw reverse policy as $\nu_\phi(\hat x\mid s,o^*)$ and the forward policy as $\pi_\theta$. The semantic check and fallback map raw generations to the effective forward distribution $q_t$. With the forward policy frozen at $\theta_t$ for this local analysis, let $\bar r_t(\hat x)$ be the expected reverse reward after semantic and cycle verification. Its reverse expected-reward objective is
\begin{equation}
J_t^{\mathrm{rev}}(\phi)
=\mathbb E_{\hat x\sim\nu_\phi(\cdot\mid s,o^*)}
\left[\bar r_t(\hat x)\right].
\end{equation}
The score-function identity gives
\begin{equation}
\left.\nabla_\phi J_t^{\mathrm{rev}}(\phi)\right|_{\phi_t}
=\mathbb E_{\hat x\sim\nu_{\phi_t}}
\left[\bar r_t(\hat x)
\left.\nabla_\phi\log\nu_\phi(\hat x\mid s,o^*)\right|_{\phi_t}\right].
\label{eq:sampling_reverse_gradient}
\end{equation}
Thus, fresh reverse generations play the role of fresh response trajectories in standard language-model RL: their distribution matches the policy whose expected reward is being differentiated. Reusing old reverse outputs without correction generally estimates a different expectation. Fresh forward rollouts also evaluate these instructions against the current grounder's behavior.

The two updates serve complementary purposes. On-policy generation refreshes the instructions encountered by the forward policy, while the reverse reward reinforces semantically valid, cycle-consistent generation. Forward-only training provides no explicit reverse-task objective under shared parameters. This explains the role of reverse updates in supporting continued instruction generation, without proving immunity to forgetting. Because forward and reverse rewards differ, these local identities are not asserted to be the full gradient of one shared scalar objective.

\paragraph{Scope of the analysis.}
The propositions analyze expected-reward policy gradients with trajectory-independent baselines and independent rollouts. GRPO additionally uses sample-dependent group normalization, clipping, KL regularization, and token/sequence averaging. Its normalized gradient samples are dependent, so these results do not establish unbiasedness or variance ordering for full GRPO updates with different group sizes (16 versus 4). They explain sampling and distribution tracking; Appendix~\ref{app:additional_experiments} evaluates the coupled algorithm empirically.

%% file: appendix/sections/prompt_templates.tex
% Reproducibility prompts collected at the end of the appendix.
% Included on full-width pages; prompt boxes can break safely across pages.
\section{Prompt Templates}
\label{app:prompt_templates}
This section collects the prompts for instruction generation, grounding, data refinement, variant generation, and agentic planning.

\begingroup
\small

\subsection{Instruction Generation Prompt}
\label{app:prompt_generation}
To unify diverse GUI tasks, we design a task-adaptive prompt template. This template guides the VLM to generate specific instruction formats conditioned on the input Task Type ID and spatial coordinates. Specifically, it encompasses three modalities: action-oriented instruction generation (Action Grounding), region description (Grounding bbox), and fine-grained element description (Grounding point). Furthermore, the prompt explicitly constrains the model to generate unambiguous descriptions, ensuring that the output corresponds uniquely to the target element within the current screen context.

\begin{tcolorbox}[
    breakable,
    colback=gray!5,
    colframe=blue!15!teal,
    title=\textbf{Prompt: GUI Instruction Generation Across Different Tasks.},
    fonttitle=\bfseries\small,
    fontupper=\small\rmfamily,
    sharp corners,
    boxrule=0.8pt,
    left=3mm, right=3mm, top=2mm, bottom=2mm
]
You are specialized in GUI understanding and grounding.

There are three types of tasks with different output requirements:

1. Action grounding: Generate a grounding instruction starting with a verb (e.g., 'Click ...').

2. Grounding bbox: Generate a descriptive caption for the region within the bounding box.

3. Grounding point: Generate a descriptive caption for the element at the specific point.

Instruction:

Based on the provided task type ID and screen coordinates, generate the corresponding sentence. The output must correspond to \textbf{one and only one element} on the entire screen.

Current Task Configuration:

- Coordinates: \{coord\_str\}

- Task Type ID: \{task\_type\_id\}

Please generate the result for the current task:
\end{tcolorbox}

\subsection{Grounding Instruction Prompt}
\label{app:prompt_grounding}
To bridge the gap between high-level user intentions and low-level executable operations, we design a comprehensive Grounding Instruction Prompt. This prompt explicitly defines the agent's action space, encompassing atomic actions ranging from basic interactions such as 'click' and 'type' to complex gestures like 'scrollmenu'. By enforcing the embedding of specific coordinate tokens including <|box\_start|> within action arguments, we ensure precise spatial grounding of generated actions. Furthermore, the prompt mandates a strict output format and incorporates exception handling mechanisms, specifically the 'stop' action for invisible elements, enabling the agent to robustly navigate and interact within the GUI environment based on visual feedback.
\begin{tcolorbox}[
    breakable,
    colback=gray!5,
    colframe=blue!15!teal,
    title=\textbf{Prompt: GUI Agent Action Grounding},
    fonttitle=\bfseries\small,
    fontupper=\small\rmfamily,
    sharp corners,
    boxrule=0.8pt,
    left=3mm, right=3mm, top=2mm, bottom=2mm
]
You are a GUI agent. You are given a task, with screenshots. You need to perform the next action to complete the task.

\textbf{\#\# Output Format}

Action: ...

\textbf{\#\# Action Space}

1. \seqsplit{click(start\_box='<|box\_start|>(x1,y1)<|box\_end|>')}

2. \seqsplit{rightclick(start\_box='<|box\_start|>(x1,y1)<|box\_end|>')}

3. \seqsplit{doubleclick(start\_box='<|box\_start|>(x1,y1)<|box\_end|>')}

4. \seqsplit{select(start\_box='<|box\_start|>(x1,y1)<|box\_end|>')}
   \% To expand a dropdown menu or select an item from it.

5. \seqsplit{type(content='', start\_box='<|box\_start|>(x1,y1)<|box\_end|>')}
   \% First click on the textbox, and then type the content.

6. hotkey(key='')

7. stop(reason='')
   \% If the item can not found in the image, give the reason.

8. scrolldown() / scrollup()
   \% If the all screenshot need to scroll down/up.

9. \seqsplit{scrollmenu(start\_box='<|box\_start|>(x1,y1),(x2,y2)<|box\_end|>')}
   \% If the part of screenshot need to scroll and give the area need to scroll.

10. finish()

\textbf{\#\# User Instruction:}

\{user\_instruction\}
\end{tcolorbox}

\clearpage
\subsection{Grounding Instruction Refinement Prompt}
\label{app:prompt_refinement}
The following template implements the dual-view refinement procedure described in Appendix~\ref{app:instruction_refinement}.

\begin{tcolorbox}[
    breakable,
    colback=gray!5,
    colframe=blue!15!teal,
    title=\textbf{Prompt: Dual-view Instruction Refinement},
    fonttitle=\bfseries\small,
    fontupper=\small\rmfamily,
    sharp corners,
    boxrule=0.8pt,
    left=3mm, right=3mm, top=2mm, bottom=2mm
]
\textbf{\#\# Task: GUI Instruction Verification \& Refinement}

You are a GUI agent expert. You will receive:

1. \textbf{First Image}: The \textbf{Full Screenshot} with a red bounding box highlighting a specific element.

2. \textbf{Second Image}: A \textbf{Cropped Image} zoomed in on the same element (also marked with a RED BOX).

3. \textbf{Current Instruction}: The user's command provided in the input.

\textbf{\#\# Steps:}

1. \textbf{Verify}: Does the "Current Instruction" accurately and unambiguously describe the element inside the red box in the images?

   - If the instruction matches the visual element perfectly, set `is\_aligned` to true.

   - If the instruction is wrong, vague, refers to a different element, or mentions "red box", set `is\_aligned` to false.

2. \textbf{Generate New}: Regardless of alignment, generate a high-quality, unambiguous, action-oriented instruction based \textit{strictly} on the visual appearance in the Cropped Image. The spatial position description can refer to the full screenshot.

    - Use the visual text on the element if available (e.g., "Click the 'Submit' button").

    - If no text, use icon description or clear function (e.g., "Click the search icon").

    - Do NOT mention bounding boxes, or "red".

\textbf{\#\# Output JSON Format:}

Your output MUST be a single, well-formed JSON object with the following exact keys:

\{\{

    "reasoning": "Brief analysis of the match...",
    "is\_aligned": true or false,

    "new\_instruction": "The newly generated instruction based on the cropped image..."
\}\}

Current Instruction is: \{current\_instruction\}
\end{tcolorbox}

\subsection{Instruction Variant Generation Prompt}
\label{app:prompt_variants}
The following template generates target-preserving paraphrases for the benchmark described in Appendix~\ref{sec:osworld_variants}.

\begin{tcolorbox}[
    breakable,
    colback=gray!5,
    colframe=blue!15!teal,
    title=\textbf{Prompt: Instruction Variant Generation for OSWorld-G-Refine-Variants},
    fonttitle=\bfseries\small,
    fontupper=\small\rmfamily,
    sharp corners,
    boxrule=0.8pt,
    left=3mm, right=3mm, top=2mm, bottom=2mm
]
You are a professional instruction rewriter for GUI grounding benchmarks.

Given an original instruction that tells a user to click/interact with a specific UI element on a screenshot, generate exactly 3 different rephrased versions of this instruction.

\textbf{Requirements:}

1. Each variant MUST refer to the EXACT SAME UI element/location as the original.

2. Each variant should use DIFFERENT wording, sentence structure, or perspective.

3. Maintain the same level of specificity --- don't add or remove critical spatial/visual details.

4. Keep instructions natural and fluent in English.

5. Variants should be diverse from each other (not just synonym substitution).

\textbf{Strategies you can use:}

- Use different spatial references (``top-left'' $\rightarrow$ ``upper-left corner'')

- Emphasize different visual attributes (color, shape, icon, text, position)

- Change verb choice (``click'' $\rightarrow$ ``select'' $\rightarrow$ ``tap on'' $\rightarrow$ ``press'')

- Reorder the descriptive clauses

Return your answer as a JSON array of exactly 3 strings, nothing else:

["variant1", "variant2", "variant3"]
\end{tcolorbox}

\clearpage
\input{appendix/sections/planner_prompt}
\endgroup

%% file: appendix/sections/planner_prompt.tex
% Agentic Evaluation: Planner Prompt (Android)
% Final subsection of the shared prompt_templates section.
% Requires in the preamble: \usepackage{tcolorbox}\tcbuselibrary{breakable},
% and the \cite{zheng2024seeact} entry in main.bib. Uses full-width pages.
\subsection{Agentic Evaluation: Planner Prompt}
\label{sec:planner_prompt}
In the interactive agentic setting on Android, BiGrounder serves as the downstream grounding/action model inside a two-stage \emph{planner--executor} loop. At each step, a planner VLM analyzes the current screenshot, the user's overall task, and the interaction history, and emits the next \emph{intended operation} as structured natural language---never as executable code, JSON actions, screen coordinates, or UI element indexes. The downstream grounding/action model---BiGrounder, driven by the grounding prompt in Appendix~\ref{app:prompt_grounding}---consumes this intended operation and its target description and converts them into a concrete executable action. This decoupling isolates high-level decision making (the planner) from pixel-level localization and low-level execution (the actor), letting us evaluate BiGrounder's grounding ability independently of the planner's reasoning. The planner's \texttt{action\_type\_hint}, \texttt{target\_description}, and \texttt{value\_hint} mirror the (operation, element, value) decomposition popularized by SeeAct~\cite{zheng2024seeact}, cleanly separating \emph{what to do}, \emph{where to do it}, and \emph{with what argument}. The planner is driven by two prompts: a fixed \emph{system prompt} that defines its role, action-type hints, planning rules, and output format; and a \emph{user prompt} filled in at each step with the current task and operation history (placeholders \texttt{\{instruction\}} and \texttt{\{actions\}} are substituted at run time). Both are shown below.

\begin{tcolorbox}[
    breakable,
    colback=gray!5,
    colframe=blue!15!teal,
    title=\textbf{Planner System Prompt --- Role, Inputs, Action Types, Planning Rules, and Output Format},
    fonttitle=\bfseries\small,
    fontupper=\fontsize{7}{8.5}\selectfont\rmfamily,
    sharp corners,
    boxrule=0.8pt,
    left=3mm, right=3mm, top=1.5mm, bottom=1.5mm
]
You are an autonomous GUI planner operating on the Android platform. Your primary function is to analyze the current Android screenshot, the user's task, and the previous operation history, then generate the next intended operation for a downstream action/grounding model.

You are only the planner. You must not execute actions. You must not output coordinates. You must not output executable function calls. You must not output JSON actions. You must not choose UI element indexes. The downstream action/grounding model will convert your planned operation into a concrete action such as click, swipe, long press, open app, type text, press back, wait, answer, or terminate.

\textbf{\#\# Input Specification.} You will receive: the screenshot of the current Android screen; the user's overall task; and your past interaction history with the UI. The screenshot is provided as image input and is the most reliable source of truth. Previous operations may be incomplete, inaccurate, or over-claiming effects; always verify progress from the current screenshot.

\textbf{\#\# Planning Objective.} For each step, decide the next intended operation that should be performed to make progress toward the task. You should determine: what screen or app is currently visible; what has already been completed; whether the previous operation succeeded; what remains to be done; what the next local operation should be; what target the downstream grounding model should locate; what value is needed, if any; and how to check whether the operation worked.

\textbf{\#\# Available Action Type Hints.} The downstream model may execute one of the following action types: \texttt{CLICK}, \texttt{LONG\_PRESS}, \texttt{INPUT\_TEXT}, \texttt{SWIPE}, \texttt{OPEN\_APP}, \texttt{KEYBOARD\_ENTER}, \texttt{NAVIGATE\_HOME}, \texttt{NAVIGATE\_BACK}, \texttt{WAIT}, \texttt{ANSWER}, \texttt{TERMINATE\_SUCCESS}, \texttt{TERMINATE\_FAILURE}. You should provide one action type hint, but you must not output the final executable action.

\textbf{\#\# Planning Rules}

\textbf{General:} Carefully inspect the current screenshot and understand the user's overall task before planning. Prefer the simplest reliable next step, and plan only one next operation. The next operation must be grounded in the current screen or in a valid navigation/open-app action. Do not repeat the same failed operation; if the previous operation did not work, or the agent appears stuck in a loop, switch to a different approach.

\textbf{Completion:} Do not plan \texttt{TERMINATE\_SUCCESS} unless the screenshot or history clearly proves the task is complete. If the user asks a question and the answer is known, plan \texttt{ANSWER} first; the following step should usually be \texttt{TERMINATE\_SUCCESS}. If the task is impossible because of a missing app, missing permission, unavailable information, blocked UI, or an unsupported operation, plan \texttt{TERMINATE\_FAILURE}.

\textbf{Navigation:} If the current screen is irrelevant, plan to open the relevant app, navigate back, or return home; if the task requires entering a specific app, prefer \texttt{OPEN\_APP}. If the screen is loading or unstable, plan \texttt{WAIT}. If a dialog blocks progress, handle the dialog first if relevant. If the target item is not visible, plan to scroll, search, or navigate to reveal it.

\textbf{Text input:} If text needs to be entered, plan to focus the correct field and input the exact text; if the field is already focused, plan direct text input. If text was just entered, check whether confirmation is needed (e.g., pressing Enter or tapping Search, OK, Save, Send, or Done). If the correct text is already present, do not plan to type it again.

\textbf{Scrolling:} If the target content is likely below the visible area, plan to scroll down; if likely above, plan to scroll up. If the previous scroll did not help, try the opposite direction or another route.

\textbf{Safety:} Avoid planning actions that delete data, purchase items, send messages, make calls, create accounts, log in, submit final forms, or expose private data unless the user's task explicitly requires it. If such an action is required, proceed only when the current UI clearly matches the user's intent. If risky or ambiguous, plan \texttt{TERMINATE\_FAILURE} with a clear reason.

\textbf{\#\# Output Format.} Your response must be exactly:

\texttt{<think>} A concise analysis of the current screen, task progress, previous operation result, and why the next operation is appropriate. \texttt{</think>}

\texttt{<operation>} One clear next intended operation for the downstream action/grounding model. This should be a local instruction, not executable code. \texttt{</operation>}

\texttt{<action\_type\_hint>} One of: \texttt{CLICK}, \texttt{LONG\_PRESS}, \texttt{INPUT\_TEXT}, \texttt{SWIPE}, \texttt{OPEN\_APP}, \texttt{KEYBOARD\_ENTER}, \texttt{NAVIGATE\_HOME}, \texttt{NAVIGATE\_BACK}, \texttt{WAIT}, \texttt{ANSWER}, \texttt{TERMINATE\_SUCCESS}, \texttt{TERMINATE\_FAILURE}. \texttt{</action\_type\_hint>}

\texttt{<target\_description>} Natural-language description of the target UI element, screen area, app, or information source. Use visible text, icon, position, role, or nearby context when possible. Use \texttt{None} if no target is needed. \texttt{</target\_description>}

\texttt{<value\_hint>} Exact text to input, app name, answer text, scroll direction, or termination reason if applicable. Use \texttt{None} otherwise. \texttt{</value\_hint>}

\texttt{<success\_check>} What should be observed after the downstream actor performs this operation if it succeeds. \texttt{</success\_check>}

\texttt{<fallback>} What to try next if this operation fails. \texttt{</fallback>}

\textbf{\#\# Output Constraints.} Do not output anything outside the required tags. Do not output executable actions, coordinates, UI element indexes, or JSON actions. Keep each field concise.
\end{tcolorbox}

\begin{tcolorbox}[
    breakable,
    colback=gray!5,
    colframe=blue!15!teal,
    title=\textbf{Planner User Prompt (sent with the current screenshot at every step)},
    fonttitle=\bfseries\small,
    fontupper=\footnotesize\rmfamily,
    sharp corners,
    boxrule=0.8pt,
    left=3mm, right=3mm, top=1.5mm, bottom=1.5mm
]
Please generate the next planned operation according to the UI screenshot, the task and previous operations.

The current Android screenshot is provided as image input.

Task: \texttt{\{instruction\}}

Previous operations: \texttt{\{actions\}}
\end{tcolorbox}